\documentclass[10pt,twocolumn,letterpaper]{article}
\usepackage{cvpr}

\usepackage{amsmath,amsfonts,bm}

\def\eqref#1{equation~\ref{#1}}

\def\1{\bm{1}}

\def\rk{{\textnormal{k}}}
\def\rl{{\textnormal{l}}}

\def\rvl{{\mathbf{l}}}

\def\rvw{{\mathbf{w}}}

\def\vg{{\bm{g}}}

\def\vl{{\bm{l}}}

\def\vv{{\bm{v}}}
\def\vw{{\bm{w}}}
\def\vx{{\bm{x}}}

\def\mB{{\bm{B}}}

\def\mG{{\bm{G}}}

\def\mX{{\bm{X}}}

\DeclareMathAlphabet{\mathsfit}{\encodingdefault}{\sfdefault}{m}{sl}
\SetMathAlphabet{\mathsfit}{bold}{\encodingdefault}{\sfdefault}{bx}{n}

\def\sA{{\mathbb{A}}}

\def\sD{{\mathbb{D}}}

\def\sK{{\mathbb{K}}}

\def\sP{{\mathbb{P}}}

\def\sS{{\mathbb{S}}}

\def\sU{{\mathbb{U}}}

\usepackage[utf8]{inputenc} 
\usepackage[T1]{fontenc}    
\usepackage{url}            
\usepackage{booktabs}       
\usepackage{amsfonts}       
\usepackage{nicefrac}       
\usepackage{microtype}      
\usepackage{xcolor}         

\usepackage{graphicx}
\usepackage{amsmath}
\usepackage{amssymb}
\usepackage{amsthm}
\usepackage{bm}
\usepackage{stfloats}
\usepackage{float}
\usepackage{multirow}
\usepackage{xspace}
\usepackage{algorithm}
\usepackage{algorithmic}
\usepackage[accsupp]{axessibility}

\usepackage[pagebackref,breaklinks,colorlinks]{hyperref}

\usepackage[capitalize]{cleveref}
\crefname{section}{Sec.}{Secs.}
\Crefname{section}{Section}{Sections}
\Crefname{table}{Table}{Tables}
\crefname{table}{Tab.}{Tabs.}

\def\cvprPaperID{4369} 
\def\confName{CVPR}
\def\confYear{2022}

\newtheorem{theorem}{Theorem}
\newtheorem{lemma}{Lemma}
\newtheorem{corollary}{Corollary}
\newtheorem{assumption}{Assumption}

\begin{document}
\title{FedCor: Correlation-Based Active Client Selection Strategy for Heterogeneous Federated Learning}

\author{Minxue Tang$^1$, Xuefei Ning$^2$, Yitu Wang$^1$, Jingwei Sun$^1$, Yu Wang$^2$, Hai Li$^1$, Yiran Chen$^1$\thanks{Corresponding Author}\\
$^1$Department of Electrical and Computer Engineering, Duke University\\
$^2$Department of Electronic Engineering, Tsinghua University\\
\tt\small $^1$\{minxue.tang,yitu.wang,jingwei.sun,hai.li,yiran.chen\}@duke.edu \\
\tt\small $^2$foxdoraame@gmail.com \quad $^2$yu-wang@tsinghua.edu.cn
}

\newcommand{\fix}{\marginpar{FIX}}
\newcommand{\new}{\marginpar{NEW}}

\maketitle

\begin{abstract}

Client-wise data heterogeneity is one of the major issues that hinder effective training in federated learning (FL). Since the data distribution on each client may vary dramatically, the client selection strategy can significantly influence the convergence rate of the FL process. Active client selection strategies are popularly proposed in recent studies. However, they neglect the loss correlations between the clients and achieve only marginal improvement compared to the uniform selection strategy. In this work, we propose FedCor---an FL framework built on a correlation-based client selection strategy, to boost the convergence rate of FL. Specifically, we first model the loss correlations between the clients with a Gaussian Process (GP). Based on the GP model, we derive a client selection strategy with a significant reduction of expected global loss in each round. Besides, we develop an efficient GP training method with a low communication overhead in the FL scenario by utilizing the covariance stationarity. Our experimental results show that compared to the state-of-the-art method, FedCorr can improve the convergence rates by $34\%\sim 99\%$ and $26\%\sim 51\%$ on FMNIST and CIFAR-10, respectively.
\end{abstract}


\section{Introduction}\label{Sec:Intro}
As a newly emerging distributed learning paradigm, federated learning (FL)~\cite{konevcny2015federated,konevcny2016federated,mcmahan2017communication,kairouz2019advances,li2020federated} has recently attracted attention because of the offered data privacy. FL aims at dealing with scenarios where training data is distributed across a number of clients. Considering limited communication bandwidth and the privacy requirement, in each communication round, FL usually selects only a fraction of clients, and the selected clients will perform multiple iterations of local updating without exposing their own datasets~\cite{mcmahan2017communication}. This special scenario also introduces other challenges that distinguish FL from the conventional distributed learning~\cite{boyd2011distributed,yang2019survey}.

One major challenge in FL is the high degree of client-wise data heterogeneity~\cite{li2020federated}, which is the inherent characteristic of a large number of clients. There have been many studies~\cite{smith2017federated,li2018federated,liang2019variance,karimireddy2019scaffold,wang2020tackling,reisizadeh2020robust,li2020lotteryfl,li2021hermes} trying to tackle non-IID (independent and identically distributed) and unbalanced data of the clients in FL. Most of these studies~\cite{li2018federated,liang2019variance,karimireddy2019scaffold,wang2020tackling} focus on amending the local model updates or the central aggregation based on FedAvg~\cite{mcmahan2017communication}. 

Recently, active client selection arises as a complement of the aforementioned studies, aiming at accelerating the convergence of FL with non-IID data. 
Some recent studies propose to assign higher probability of being selected to the clients with larger training loss value~\cite{goetz2019active,cho2020client}. 
However, they neglect the correlations between the clients and consider their losses independently, which leads to only marginal performance improvement. In this paper, we propose a correlation-based active client selection strategy that can effectively alleviate the accuracy degradation caused by data heterogeneity and significantly boost the convergence of FL. Our key idea is mainly based on the following intuitions:

\begin{enumerate}
    \item Clients do not contribute \textbf{equivalently}. 
    For example, training with a large and balanced dataset on a ``good'' client can reduce the losses of most clients, while training with a small and extremely biased dataset on a ``bad'' client may increase the losses of other clients.
    \item Clients do not contribute \textbf{independently}. 
    The influence of selecting one client depends on the other selected clients because their local updates will be aggregated. 
\end{enumerate}

\begin{figure}[t]
  \centering
    \includegraphics[width=0.95\linewidth]{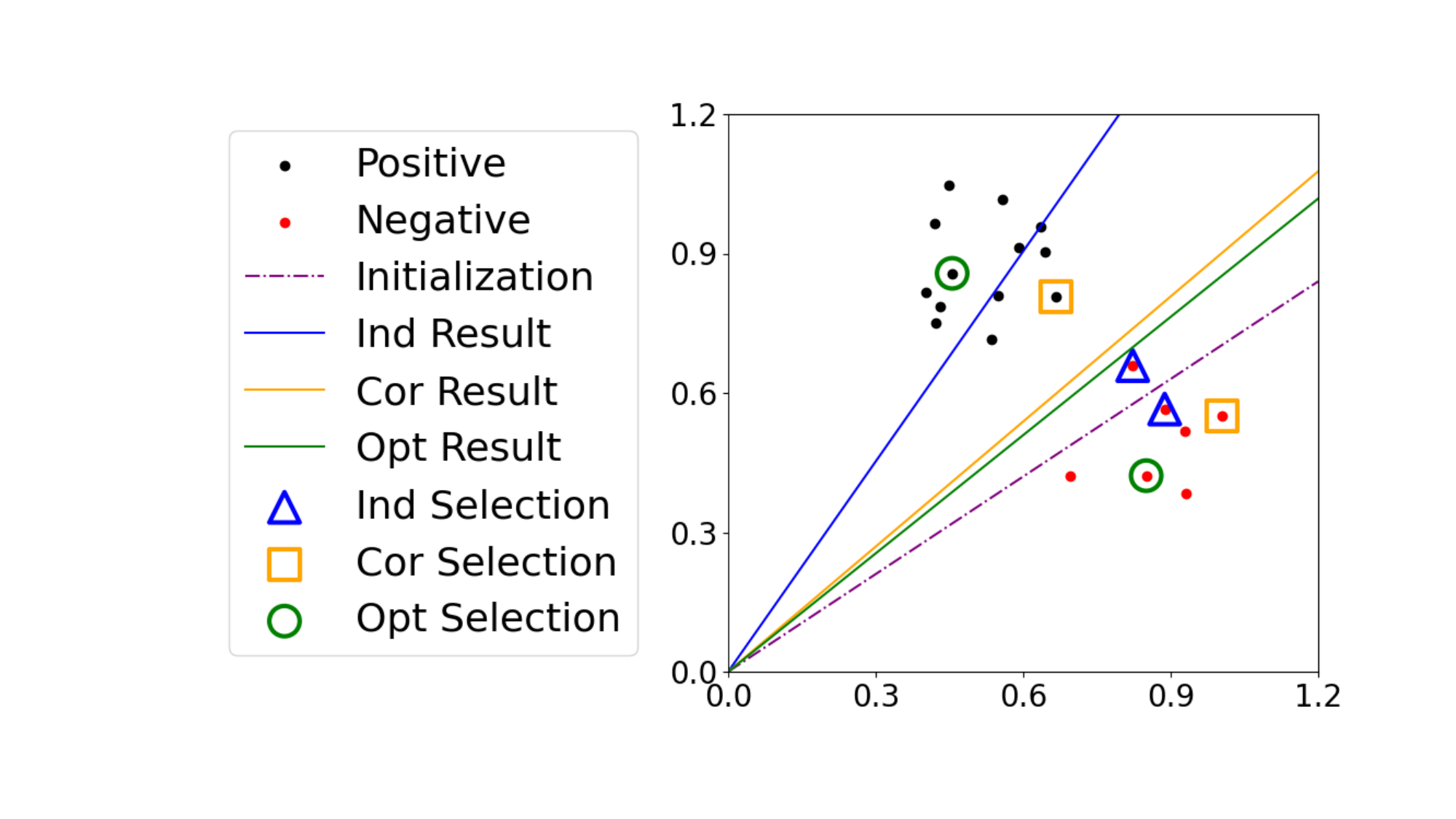}
  \caption{A toy experiment of different client selection strategies. }
\label{fig:toy}
\end{figure}
A toy experiment shown in \cref{fig:toy} also illustrates the necessity of considering the correlations for client selection. In this  experiment, each client has only one data sample, and thus each data point in the figure represents a client. The task is to select two clients (different markers represent the client selections of different strategies) for training a binary classifier (shown as the lines). The selection strategy that independently selects two clients with the highest local losses (``Ind Result'') fails to reduce the global loss. In contrast, our method considers the correlations between the clients (``Cor Result'') and derives a client selection that can achieve an almost lowest global loss (``Opt Result'').

Based on the above intuitions, this work proposes FedCor, an FL framework built on a correlation-based client selection strategy, to boost the convergence of FL. Our main contributions are summarized as follows:
\begin{enumerate}
    \item We model the client loss changes with a Gaussian Process (GP) and propose an interpretable client selection strategy with a significant reduction of the expected global loss in each communication round.
    
    \item We propose a GP training method that utilizes the covariance stationarity to reduce the communication cost. Experiments show that the GP trained with our method can capture the client correlations well.
    \item Experimental results demonstrate that FedCor stabilizes the training convergence and significantly improves the convergence rates by $34\%\sim 99\%$ and $26\%\sim 51\%$ on FMNIST and CIFAR-10, respectively.
\end{enumerate}

 
\section{Related Work}\label{Sec:Rel}

An important characteristic of FL~\cite{konevcny2015federated,konevcny2016federated,mcmahan2017communication} is the heterogeneity of clients, which raises new challenges of the training~\cite{kairouz2019advances,li2020federated,wang2021field}. There are two kinds of heterogeneity in FL: systemic heterogeneity (computation ability, communication bandwidth, etc.) and statistical heterogeneity (non-IID, imbalanced data distribution)~\cite{li2018federated,li2020federated}. This work mainly focuses on the latter one. A number of methods have been proposed to improve the basic FL algorithm, FedAvg~\cite{mcmahan2017communication}, in heterogeneous settings. 
Some of them manipulate the local training loss like adding regularization terms to stabilized the  training~\cite{li2018federated,shoham2019overcoming,liang2019variance,karimireddy2019scaffold,hsu2020federated}, while some other works amend the aggregation method to reduce the variance~\cite{wang2020tackling,murata2021bias}. 

Complementary to such methods, another way to improve the convergence of FL in non-IID settings is active client selection, which tries to strategically select clients for training in each round in stead of uniformly selecting. 
Goetz et al.~\cite{goetz2019active} first propose to assign a high selection probability to the clients with large local loss. Cho et al.~\cite{cho2020client} select $C$ clients with the largest loss among a randomly sampled subset $\sA\subseteq\sU$ with size $d>C$ to reduce the selection bias.
However, neither of them consider the correlations between clients while making the client selection. 
\section{Preliminary}\label{Sec:pre}
FL seeks for a global model $\vw$ that achieves the best performance (e.g., the highest classification accuracy) on all $N$ clients. The global loss function in FL is defined as:
\begin{align}
    L(\vw) &= \sum_{k=1}^N \frac{|\sD_k|}{\sum_j|\sD_j|}l(\vw;\sD_k)=\sum_{k=1}^N p_kl_k(\vw),\label{eq:global_loss}\\
    l_k(\vw)&=l(\vw;\sD_k)=\frac{1}{|\sD_k|}\sum_{\mathbf{\xi}\in \sD_k}l(\vw;\mathbf{\xi}),
\end{align}
where $l(\vw;\mathbf{\xi})$ is the objective loss of data sample $\mathbf{\xi}$ evaluated on model $\vw$. We refer to $l_k(\vw)$ as the local loss of client $k$, which is evaluated with the local dataset $\sD_k$ (of size $|\sD_k|$) on client $k$. The weight $p_k = |\sD_k|/\sum_j |\sD_j|$ of the client $k$ is proportional to the size of its local dataset.


In consideration of the privacy and communication constraints, FL algorithms usually assume partial client participation and perform local model updates.  In particular, in communication round $t$, only a subset $\sK_t$ with size $|\sK_t|=C\le N$ of the overall client set $\sU$ is selected to receive the global model $\vw^{t}$ and conduct training with their local dataset for several iterations independently. After the local training, the server collects the trained models from these selected clients and aggregates them (usually by averaging~\cite{mcmahan2017communication}) to produce a new global model $\vw^{t+1}$. We formulate this procedure as follows:
\begin{align}
    \vw_k^{t+1}&=\vw^{t}-\eta_t\tilde\nabla l_k(\vw^{t}),\\
    \vw^{t+1}(\sK_t)&=\frac{1}{C}\sum_{k\in \sK_t}\vw_k^{t+1}\\
    &=\vw^t-\frac{\eta_t}{C}\sum_{k\in \sK_t}\tilde\nabla l_k(\vw^t)\label{eq:fl_update},
\end{align}
where $\eta_t$ is the learning rate and $\tilde\nabla l_k(\vw^t)$ is the equivalent cumulative gradient~\cite{wang2020tackling} in the $t$-th communication round. More specifically, for an arbitrary optimizer on the client $k$, it produces $\Delta \vw_k^{t,\tau}=-\eta \boldsymbol d^{t,\tau}_k$ as the local model update at the $\tau$-th iteration in this round, and the cumulative gradient is calculated as $\tilde\nabla l_k(\vw^{t})=\sum_\tau \boldsymbol d_k^{t,\tau}$.

\section{Methodology}\label{Sec:Method}
In this section, we elaborate our proposed method, i.e., FedCor, that can effectively boost the convergence of FL. We first formulate our goal of accelerating the convergence of FL as optimization problems that maximize the posterior expectation of loss decrease in \cref{SubSec:Pre}. Then, \cref{SubSec:GP} demonstrates empirical evidence that the prior distribution of loss changes in each communication round can be modeled as Gaussian Processes (GP). Based on this observation, we utilize GP to solve the optimization problems and obtain an effective client selection strategy for heterogeneous FL in \cref{SubSec:Sel}. We further analyze the selection criterion of our client selection strategy and give out its intuitive interpretation in \cref{SubSec:Insight}. Finally, in \cref{SubSec:Train}, we describe how we train the GP parameters in communication-constrained FL.

\subsection{Problem Formulation}\label{SubSec:Pre}
To achieve a fast convergence, we hope to find the client selection strategy which can lead to the maximal global loss decrease after each communication round. 
Accordingly, we define our target as solving a series of optimization problems, one for each communication round $t$:
\begin{equation}\label{eq:opt}
\begin{aligned}
    \min_{\sK_t}\quad &\Delta L^{t}(\sK_t)=L(\vw^{t+1}(\sK_t))-L(\vw^t)\\
    \text{subject to }\quad& \vw^{t+1}(\sK_t)=\vw^t-\frac{\eta_t}{C}\sum_{k\in \sK_t}\tilde\nabla l_k(\vw^t).
\end{aligned}
\end{equation}

It is impractical in FL to search for the best client selection with multiple trials of different client selections since it introduces large communication and computation overhead. Therefore, we need an efficient way to predict the global loss decreases for different client selections and make a decision with very limited trials. To achieve this goal, we first reformulate the optimization problem in \cref{eq:opt} with the following lemma. The proof of this lemma is in \cref{Apx:lemma2_proof}.

\begin{lemma}\label{lemma2}
The optimization problem in \cref{eq:opt} is approximately equivalent to the following probabilistic form.
\begin{equation}\label{eq:gp_objective}
    \min_{\sK_{t}}\quad\mathbb{E}_{\Delta \rvl^t|\Delta\bm l^{t}_{\sK_{t}}(\sK_t)}\Big[\sum_i p_i\Delta l^{t}_i\Big] = 
    \sum_i p_i\tilde{\mu}^{t}_i(\Delta  \bm l^{t}_{\sK_{t}}(\sK_t)),
\end{equation}
where $\Delta \rvl^t=[\Delta\rl^t_1,\cdots,\Delta\rl^t_N]$ is the loss changes of all clients in round $t$, which is a random variable w.r.t random client selection in round $t$. $\tilde{\bm{\mu}}^{t}(\Delta  \bm l^{t}_{\sK_{t}}(\sK_t))$ is the posterior mean of $\Delta\rvl^t$ conditioned on $\Delta  \bm l^{t}_{\sK_{t}}(\sK_t)=[\Delta l^t_i(\sK_t)]_{i\in \sK_t}$.
\end{lemma}

The reformulated objective in \cref{eq:gp_objective} tells that if we can predict the loss changes of those clients selected for training ($\Delta\bm l^{t}_{\sK_{t}}(\sK_t)$), we can predict the global loss change with its posterior mean and make decision according to it. Now what we need is a probabilistic model of the loss changes $\Delta \rvl^t$ to make the prediction and calculate the posterior.

\subsection{Modeling Loss Changes with GP}\label{SubSec:GP}
\begin{figure}[t]
\centering
\includegraphics[width=1.0\linewidth]{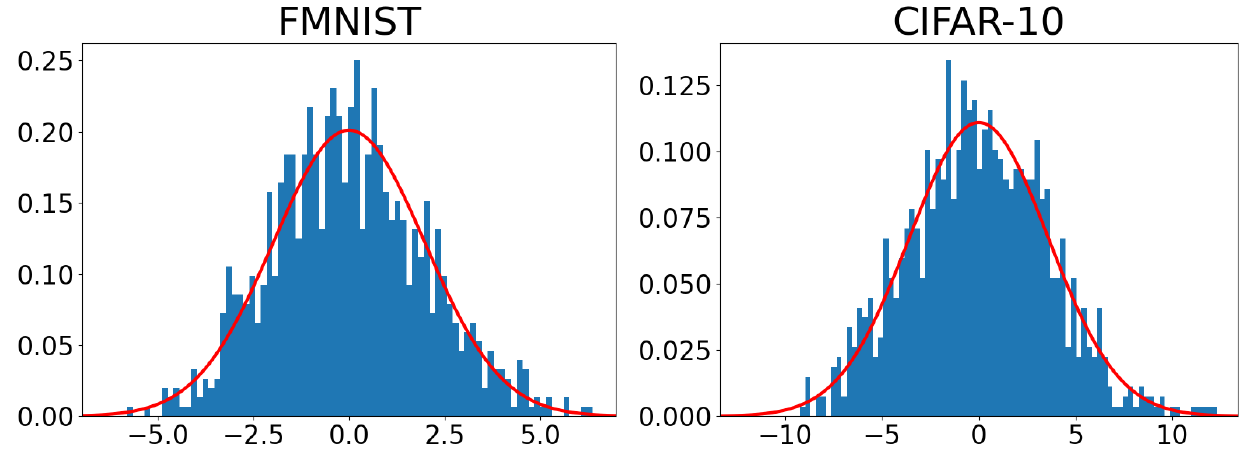}
\caption{
Histograms of the first principle component in Non-IID FL~\cite{mcmahan2017communication}. More details and full results can be found in \cref{Apx:mvnt}.}
\label{fig:mvnt_simp}
\end{figure}

It is a common practice to assume a GP prior over an unknown objective function in Bayesian Optimization~\cite{brochu2010tutorial,vien2018bayesian}. Our preliminary investigation (partly) shown in \cref{fig:mvnt_simp} also indicates that the prior distribution of the loss changes in one communication round follow a GP. Specifically, we randomly sample a number of client selections and perform one round of training to get samples of the loss changes. Then, we conduct PCA on these loss change samples and plot histograms of the first several principle components. The red line in the \cref{fig:mvnt_simp} is the Gaussian PDF with the sample mean and sample variance. And we can see that this Gaussian distribution can approximate the distribution of the samples well. A mathematical explanation of this observation is also given out in \cref{Apx:gp_analysis}.




Accordingly, we propose to model the loss changes in one communication round $t$ with a GP prior as follows: 

\begin{equation}\label{eq:GP_all}
\Delta \rvl^t=[\Delta\rl^t_1,\cdots,\Delta\rl^t_N]\sim\mathcal{N}(\Delta \bm l^t;\bm{\mu}^t,\bm{\Sigma}^t).
\end{equation} 

\noindent\textbf{Remark.} In order to efficiently learn the covariance in FL, rather than directly working with the covariance matrix, we embed all clients into a continuous vector space and use a kernel function to calculate the covariance (see \cref{SubSec:Train}).
Thus, we still use the term GP instead of Multivariate Gaussian Distribution, though the dimension of $\Delta \rvl^t$ is finite. 

A good property of GP is that we can get a closed form of the posterior expectation in \cref{eq:gp_objective}, which makes our client selection strategy interpretable. In the next sections, we will propose our client selection strategy based on the GP model, and then give an interpretation of it. We leave the training method for the parameters ($\bm{\mu}^t,\bm{\Sigma}^t$) in GP to \cref{SubSec:Train}.


\subsection{Client Selection Strategy}\label{SubSec:Sel}

While we have get the probabilistic model to calculate the posterior expectation, it is still not determined how to predict the loss changes of the clients selected for training, namely $\Delta\bm l^{t}_{\sK_{t}}(\sK_t)$. Inspired by UCB methods\cite{cox1992statistical,auer2002using,srinivas2012information}, we develop an iterative method that predict the loss change and select one client in each iteration, as shown in \cref{alg:selection}. 
There are three steps in one iteration:

\begin{algorithm}[tb]
\caption{Client Selection Strategy with GP}
\label{alg:selection}
\begin{algorithmic}[1]
\REQUIRE $\bm{\mu}^t$ and $\bm{\Sigma}^t$ of the GP, scale factor $\bm\alpha^t$
\ENSURE Client Selection $\sK_t$
\STATE Initialize $\sK_t\leftarrow \emptyset$, $\sP\leftarrow \sU$.
\WHILE{$|\sK_t|<C$}
    \FOR{each client $k\in \sP$}
        \STATE Predict its loss change if select it: $\Delta\hat l_k^t=\mu_k^t-\alpha_k^t\sigma_k^t$.
        \STATE Calculate the posterior mean of the loss changes $\tilde{\bm{\mu}}^t(\Delta \hat l_k^t)$.
    \ENDFOR
\STATE Select the client by $k^*=\mathop{\arg\min}_k\sum_i p_i\tilde{\mu}^{t}_i(\Delta  \hat l^{t}_k)$.
\STATE Add $k^*$ into $\sK_t$ and remove it from $\sP$.
\STATE $\bm{\mu}^t\leftarrow\tilde{\bm{\mu}}^t(\Delta \hat l_{k^*}^t),\bm{\Sigma}^t\leftarrow\tilde{\bm{\Sigma}}^t(\Delta \hat l_{k^*}^t)$.
\ENDWHILE
\end{algorithmic}
\end{algorithm}

\noindent\textbf{(i) Prediction.} 
In each iteration, we first make an prediction $\Delta \hat l^{t}_k$ for each client $k$ if it is selected. Generally, the selected client would have a large loss decrease since it directly participate in the model update. Thus, we propose to use the lower confidence bound as the prediction:
\begin{align}\label{eq:guess} 
\Delta\hat l_k^t&=\mu_k^t-\alpha_k^t\sigma_k^t;\quad 
\alpha_k^t=a\beta^{\tau_k^t},
\end{align}
where $\sigma_k^t = \sqrt{\Sigma^t_{k,k}}$, and $a$ is a scale constant. $\beta\in(0,1)$ is an annealing coefficient, and its index $\tau_k^t$ denotes how many times client $k$ has been selected. We will discuss this annealing coefficient more in \cref{SubSec:Train}. 

\noindent\textbf{(ii) Selection.}
The client $k^*$ is selected to minimize the posterior expectation of the overall loss conditioned on its loss change prediction made in the last step:
\begin{equation}\label{eq:selection}
    k^*=\mathop{\arg\min}_k\sum_i p_i\tilde{\mu}^{t}_i(\Delta\hat l_k^t)
\end{equation}



\noindent\textbf{(iii) Posterior.}
After selecting the client $k^*$, we update the GP for the next iteration with the posterior conditioned on the loss change prediction of $k^*$:
\begin{equation}
    \bm{\mu}^t\leftarrow\tilde{\bm{\mu}}^t(\Delta \hat l_{k^*}^t),\quad\bm{\Sigma}^t\leftarrow\tilde{\bm{\Sigma}}^t(\Delta \hat l_{k^*}^t).
\end{equation}
By updating the GP with its posterior, we iteratively add conditions into the probabilistic model to approach the fully conditioned distribution $p(\Delta \bm l^t|\Delta\bm l^{t}_{\sK_{t}}(\sK_t))$, and make the next prediction of the loss change more accurate.

There are some similarities between our method and traditional Bayesian Optimization: Using GP as a prior of the objective function, and using UCB as well as posterior distribution for iterative selection~\cite{cox1992statistical,brochu2010tutorial,srinivas2012information}. However, there is a key difference: In each communication round, we determine the client selection with only predictions instead of measurements of the global loss changes, while traditional Bayesian Optimization requires a sequence of measurements as new information to make decisions. The measurements of global loss changes will introduce large communication overhead and are unfeasible in FL. 


\begin{algorithm}[tb]
\caption{FedCor}
\label{alg:whole}
\begin{algorithmic}[1]
\STATE Initialize $\mX_0$ and Global Model $\vw_0$.
\FOR{each round $t=0,1,...$}
    \IF{$t\%\Delta t$==0}
        \STATE Uniformly sample $S$ client selections $\sS_{t,i},i=1,2,...,S$.
        \FOR{$i=1,2,...,S$}
            \STATE $\vw^{t+1}(\sS_{t,i})\leftarrow\vw^t-\frac{\eta_t}{C}\sum_{k\in \sS_{t,i}}\tilde\nabla l_k(\vw^t)$.
            \STATE Collect $\Delta \bm l^{t}(\sS_{t,i})\leftarrow \bm l(\vw^{t+1}(\sS_{t,i}))-\vl(\vw^t)$.
        \ENDFOR
        \STATE Reset $\alpha_k\leftarrow 1,\forall k\in \sU$.
    \ENDIF
    \STATE Update $\mX_t$ with \cref{eq:train}.
    \STATE Select clients $\sK_t$ with \cref{alg:selection} ($\bm{\mu}^t=\bm 0, \bm{\Sigma}^t = {\mX^t}^T\mX^t, \bm\alpha^t=\bm\alpha$).
    \STATE $\vw^{t+1}\leftarrow\vw^{t+1}(\sK_t)=\vw^t-\frac{\eta_t}{C}\sum_{k\in \sK_t}\tilde\nabla l_k(\vw^t)$.
    \STATE Update $\bm\alpha_{\sK_t}\leftarrow\beta\bm\alpha_{\sK_t}$.
\ENDFOR
\end{algorithmic}
\end{algorithm}

\subsection{Insights into Our Selection Strategy}\label{SubSec:Insight}
In this section, we give an intuitive interpretation of our selection strategy and show the benefits of it within a simple case. A more detailed analysis of the selection criterion and convergence of FedCor can be found in \cref{Apx:convergence}.  

For simplicity, we omit all superscript $t$ in this section. \cref{lemma3} gives the selection criterion of FedCor in a simple case where we only select two clients, and the proof can be found in \cref{Apx:lemma3_proof}.

\begin{lemma}\label{lemma3}
The selection criterion of FedCor when selecting two clients $k_1$ and $k_2$ can be written as
\begin{align}
    k_1=\mathop{\arg\max}_k\quad &\beta^{\tau_k}\sum_ip_i\sigma_ir_{ik},\label{eq:sim_selection}\\
    k_2=\mathop{\arg\max}_{k'}\quad &\frac{\beta^{\tau_{k'}}\Big[\overbrace{\sum_ip_i \sigma_{i}r_{ik'}}^{\text{(A)}}-r_{k_1k'}\overbrace{\sum_ip_i \sigma_ir_{ik_1}}^{\text{(B)}}\Big]}{\sqrt{1-r_{k'k_1}^2}},\label{eq:post_criterion}
\end{align}
where $r_{ij}=\Sigma_{i,j}/\sigma_i\sigma_j$ is the Pearson correlation coefficient.
\end{lemma}

\noindent\textbf{(i) Single-Iteration.} 
\cref{eq:sim_selection} has a clear interpretation to select the client that has large correlations with other clients ($r_{ik}$),
so that other clients can benefit more from training on the selected client.
Our selection criterion takes the correlations between the clients into consideration, and can conduct better selection compared with those algorithms that only consider the loss of each client independently~\cite{goetz2019active,cho2020client}.

\noindent\textbf{(ii) Multi-Iteration.}
In \cref{eq:post_criterion}, term (A) and (B) are the single-iteration selection criterion in \cref{eq:sim_selection} of client $k'$ and $k_1$, respectively. Since we have maximized (B) when selecting client $k_1$, term (B) is usually positive. Therefore,
the selection of $k'$ does not only consider its correlations with other clients ($r_{ik'}$), but also prefers the clients that have small correlations $r_{k_1k'}$ with the previous selected client $k_1$.
This criterion penalizes selection redundancy and leads to a client selection with diverse data, which reduces the variance and makes the training process more stable. Since clients with similar data generate similar local updates, selecting redundant clients only brings marginal gains to the global performance or would even drive the optimization into bad local optimum. 
This selection preference is also demonstrated in \cref{fig:toy}, 
where FedCor chooses one positive and one negative point as the optimal selection does.

\subsection{Training GP in FL}\label{SubSec:Train}
As a classical machine learning model, GP has been widely discussed and well studied~\cite{williams2006gaussian}. There have been many methods to train the parameters in GP, namely, the covariance $\bm{\Sigma}^t$ in \cref{eq:GP_all}
\footnote{We do not train $\bm{\mu}$ and set it to $\bm 0$, since it does not affect the selection strategy as we can see in \cref{lemma3} and \cref{Apx:convergence}.}. Nevertheless, to make the GP training feasible in the communication-constrained FL procedure, we should revise the GP training method to reduce the number of samples and better utilize historical information. 

In GP, a kernel function $K(\bm x_i,\bm x_j)$ is used to calculate the covariance~\cite{williams2006gaussian} as $\Sigma^t_{i,j} = K(\bm x_i^t,\bm x_j^t)$,
where $\bm x_i^t,\bm x_j^t$ are the features of the data points $i$ and $j$, respectively. 
Following this, we assign a trainable embedding in a latent space to each client. The embedding of the $k$-th client is noted as $\bm x_k^t\in\mathbb{R}^d$ ($d < N$), and we choose the kernel function as
\begin{equation}
    K(\bm x_i^t,\bm x_j^t) = {\bm x_i^t}^T\bm x_j^t,
\end{equation}
which is a homogeneous linear kernel~\cite{williams2006gaussian}. This low-rank formulation reduces the number of parameters we need to learn, thus making the GP training more data-efficient.


A commonly used GP training method is maximum likelihood evaluation, where we uniformly sample $S$ client selection $\{\sS_{t,i}:i=1,\cdots,S\}$, and maximize the likelihood of the corresponding loss changes $\{\Delta \bm l^{t}(\sS_{t,i}):i=1,\cdots,S\}$ to learn the embedding matrix $\mX^t=[\vx_1^t,\cdots,\vx_N^t]$:
\begin{equation}\label{eq:ori_gp_train}
\mX^t = \mathop{\arg\max}_{\mX} \sum_{i=1}^S\log p(\Delta \bm l^{t}(\sS_{t,i})|\mX).
\end{equation}
However, to collect each sample $\Delta \bm l^{t}(\sS_{t,i})$, we have to broadcast $\vw^{t+1}(\sS_{t,i})$ to all the clients. And since a large $S$ is usually required for an unbiased estimation in each communication round $t$, the vanilla training procedure in \cref{eq:ori_gp_train} introduces a high communication overhead.

\begin{figure}[t]
\centering
\includegraphics[width=1.0\linewidth]{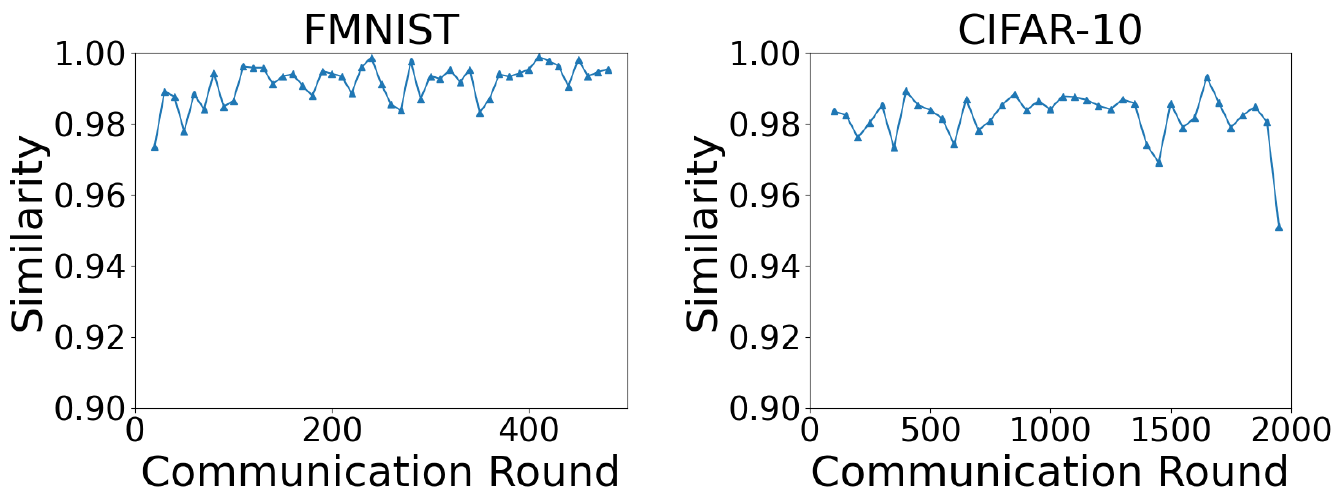}
\caption{Covariance Stationarity in Non-IID FL~\cite{mcmahan2017communication}. Full experiment results and more details can be found in \cref{Apx:cov_stat}.}
\label{fig:sim_simp}
\end{figure}

Actually, the correlations between loss changes of different clients mainly arise from similarities between their datasets, which are invariant during the FL process. Thus, we hypothesise that the covariance also changes slowly in the concerned time range. To verify this, we use a large number of samples to evaluate the covariance $\bm{\Sigma}^t$ in each communication round, and calculate the cosine similarity between $\bm{\Sigma}^t$ and $\bm{\Sigma}^{t+\Delta t}$. We set $\Delta t=10$ for FMNIST and $\Delta t=50$ for CIFAR-10. As shown in \cref{fig:sim_simp}, we can see that the similarity keeps very high ($>0.97$ for FMNIST and $>0.95$ for CIFAR-10) during the whole FL training process. 

Accordingly, we do not need to update $\mX^t$ in every round but inherit the embedding matrix $\mX^{t-1}$ from the last round and train it only every $\Delta t$ rounds. Furthermore, we can reuse historical samples for GP training to reduce the number of samples $S$ that we need to collect in each GP training round. We summarize our update rule of $\mX^t$ as follows:
\begin{equation}\label{eq:train}
\begin{aligned}
    \mX^t = 
    \begin{cases}
    \mX^{t-1}, \quad t\%\Delta t\neq0;\\
    \mathop{\arg\max}_\mX \Phi_t(\mX),\quad t\%\Delta t=0,
    \end{cases}
\end{aligned}
\end{equation}
where
\begin{align}
    \Phi_t(\mX)=\sum_{m=0}^M\sum_{i=1}^S\gamma^m\log  p(\Delta \bm l^{t-m\Delta t}(\sS_{t-m\Delta t,i})|\mX).
\end{align}
$M$ is the number of reused historical samples, and $\gamma<1$ is the discount factor to weight the historical samples. 
Our method is able to reduce the communication overhead with a large $\Delta t$ and $S=1$, while guaranteeing the performance.

As we only update the covariance $\bm{\Sigma}$ every $\Delta t$ rounds, the annealing factor $\beta^{\tau_k}$ can prevent us from making the same selection during the $\Delta t$ rounds. Repeatedly training with the same group of clients would cause the global model to overfit on their data, which may hinder the convergence of FL. In practice, we reset $\tau_k$ to $0$ after each GP training round to achieve the fastest convergence while avoiding overfitting on some clients.

We summarize our overall framework FedCor in \cref{alg:whole}. It is noteworthy that our method is orthogonal to existing FL optimizers that amend the training loss or the aggregation scheme, e.g., FedAvg~\cite{mcmahan2017communication} and FedProx~\cite{li2018federated}. So our method can be combined with any of them. 

\section{Experiments}\label{Sec:Exp}

\begin{figure*}[tb]
\centering
\includegraphics[width=0.75\linewidth]{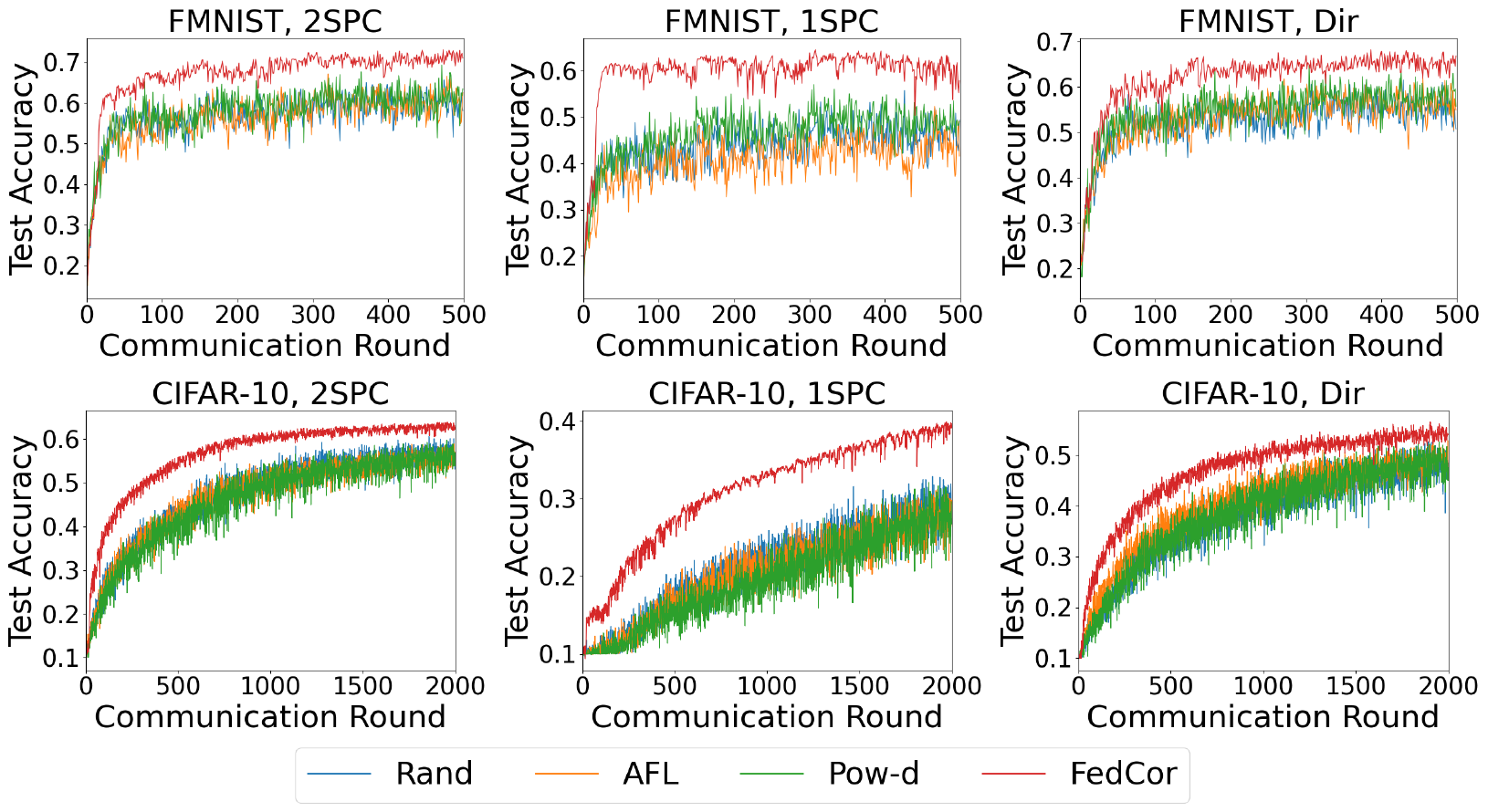}
\caption{Test accuracy on FMNIST and CIFAR-10 under three heterogeneous settings (2SPC, 1SPC and Dir). All experiments in one figure share the same hyperparameters except for the client selection strategy. }
\label{fig:convergence}
\end{figure*}

\begin{table*}[tb]
\centering
\resizebox{\textwidth}{!}{
\begin{tabular}{cccc|cccc}
\hline
 \multirow{2}{*}{Method}
&\multicolumn{3}{c|}{FMNIST}&\multicolumn{3}{c}{CIFAR-10}\\
\cline{2-7}
       & 2SPC($69\%$)      & 1SPC($62\%$)       & Dir($64\%$)        &  2SPC($62\%$)        &    1SPC($36\%$)  & Dir($54\%$) \\ 
\hline
Rand   & $295.8\pm92.0$    & N/A                & $141.0\pm73.0$     & $1561.2\pm236.2$     & $1750.4\pm190.3$     & N/A \\
AFL    & $218.6\pm117.3$   & N/A                & $169.0\pm166.1$    &      N/A    & $1845.2\pm28.8$      & $1524.4\pm267.9$ \\
Pow-d  & $126.6\pm78.2$    & $167.2\pm72.3$     & $123.0\pm101.0$    & $1558.2\pm227.0$      & $1752.2\pm186.2$     & $1355.2\pm151.3$ \\\hline
FedCor (Ours)  & $\bm{94.8\pm18.4}$& $\bm{84.0\pm53.1}$ & $\bm{68.8\pm27.5}$ &$\bm{1033.4\pm123.7}$ & $\bm{1269.2\pm70.6}$ &$\bm{1076.8\pm262.8}$ \\
\hline
\end{tabular}}
\caption{The number of communication rounds for each selection strategy
to achieve target test accuracies (specified in parentheses) under three heterogeneous settings (2SPC, 1SPC and Dir). The results consist of the mean and the standard deviation over 5 random seeds. N/A means that the corresponding selection strategy cannot achieve the target accuracy with some random seeds within the maximal number of communication rounds ($500$ for FMNIST and $2000$ for CIFAR-10).}
\label{tab:rounds}
\end{table*}
\subsection{Experiment Settings}\label{Sec:exp_settings}
We conduct experiments on two datasets, FMNIST~\cite{xiao2017fashion} and CIFAR-10~\cite{krizhevsky2009learning}. For FMNIST, we adopt an MLP model with two hidden layers, and this model achieves an accuracy of $85.92\%$ with centralized training. For CIFAR-10, we adopt a CNN model with three convolutional layers followed by one fully connected layer, and this model can achieve an accuracy of $73.84\%$ with centralized training. More details on the model construction and training hyperparameters can be found in \cref{Apx:Model}. 
For each dataset, we experiment with three different heterogeneous data partitions on $N=100$ clients as follows.

    \noindent\textbf{(i) 2 shards per client (2SPC)}: This setting is the same as the non-IID setting in \cite{mcmahan2017communication}. We sort the data by their labels and divide them into $200$ shards so that all the data in one shard share the same label. We randomly allocate these shards to clients, and each client has two shards. Since all the shards have the same size, the data partition is balanced. That is to say, all the clients have the same dataset size. We select $C=5$ clients in each round within this setting.
    
    \noindent\textbf{(ii) 1 shard per client (1SPC)}: This setting is similar to the 2SPC setting, and the only difference is that each client only has one shard, i.e., each client only has the data of one label. This is the data partition with the highest heterogeneity, and it is also balanced. We select $C=10$ clients in each round within this setting.
    
    \noindent\textbf{(iii) Dirichlet Distribution with $\alpha=0.2$ (Dir)}: We inherit and slightly change the setting from \cite{hsu2019measuring} to create an unbalanced data partition. We sample the ratio of the data with each label on one client from a Dirichlet Distribution parameterized by the concentration parameter $\alpha=0.2$. More details 
    can be found in the Appendix~\ref{Apx:Dirichlet}. We select $C=5$ clients in each round within this setting.

We divide the training process of FedCor into two phases:
    (i) Warm-up phase: We uniformly sample client selection $\sK_t$ and collect the loss values of all the clients in $\sU$ to train the GP in each round, i.e., $\Delta t=1$ and $S=1$. We set the length of the warm-up phase to $15$ for FMNIST and $20$ for CIFAR-10. 
(ii) Normal phase: After the warm-up phase, we follow Algorithm~\ref{alg:whole} to select clients and update the GP. 

In all the experiments, we use FedAvg~\cite{mcmahan2017communication} as the FL optimizer. We present the average results using five random seeds in all experiments. We will first show that our method can achieve faster and more stable convergence, compared with three baselines: random selection (Rand), Active FL (AFL)~\cite{goetz2019active} and Power-of-choice Selection Strategy (Pow-d)~\cite{cho2020client}. Then, we will give ablation studies on the GP training interval $\Delta t$ as well as the annealing coefficient $\beta$. Finally, we visualize the client embeddings $\mX$ with t-SNE~\cite{maaten2008visualizing} and show that FedCor can effectively capture the correlations.

\begin{figure*}[tb]
\centering
\includegraphics[width=0.75\linewidth]{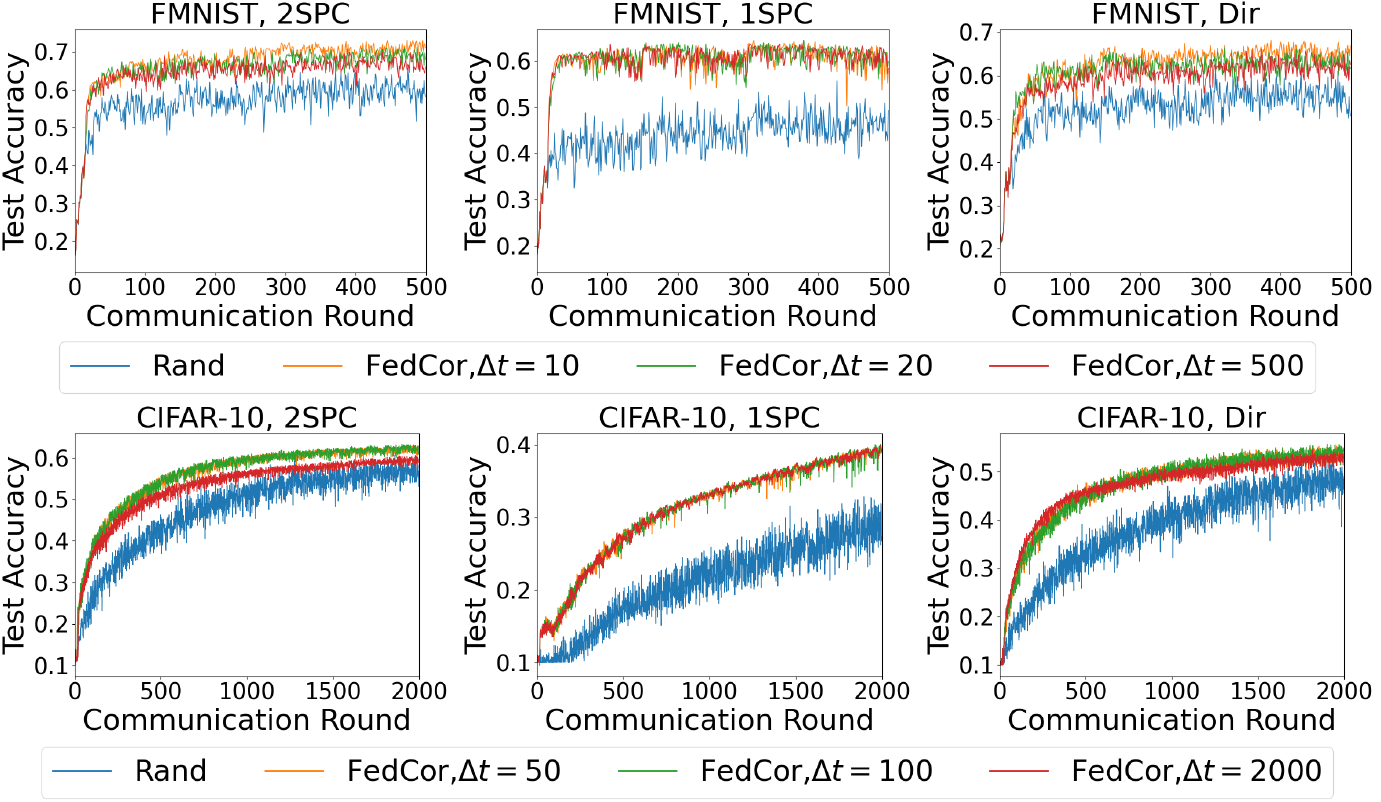}
\caption{Test accuracy with different GP training interval $\Delta t$ on FMNIST and CIFAR-10 under 2SPC, 1SPC and Dir.}
\label{fig:interval}
\end{figure*}

\subsection{Convergence under Heterogeneous Settings}

We compare the convergence rate of our method FedCor with the other baselines on both FMNIST and CIFAR-10, and demonstrate the results in Figure~\ref{fig:convergence}. We set the GP update interval $\Delta t=10$ and the annealing coefficient $\beta=0.95$ for FMNIST experiments, and $\Delta t=50$ and $\beta=0.9$ for CIFAR-10 experiments. 

As shown in Figure~\ref{fig:convergence}, FedCor achieves the highest test accuracy and the fastest convergence in all experiments. While other active client selection strategies show only slight or even no superiority compared with the fully random strategy, our method clearly outperforms all baselines, especially under the extremely heterogeneous setting when data on each client contains only one label (1SPC). Furthermore, the learning curves of FedCor are more smooth and less noisy than those of other methods, meaning that FedCor reduces the variance and makes the federated optimization more stable.

Table~\ref{tab:rounds} shows the numbers of communication rounds for each selection strategy to achieve a specified test accuracy. We can see that FedCor achieves the specified accuracy $34\%\sim 99\%$ and $26\%\sim 51\%$ faster than Pow-d on FMNIST and CIFAR-10, respectively.


\subsection{Results with Larger GP Training Interval}\label{SubSec:Int}

Collecting training data in the GP update rounds brings communication overhead, since we need to broadcast the model to all the clients. Thus, it is important to investigate the minimal GP update frequency. We vary the GP training interval and show the accuracy curves in Figure~\ref{fig:interval}. We set $\Delta t= 10,20,500$ with $\beta=0.95,0.95,0.99$ for the experiments on FMNIST, and $\Delta t=50,100,2000$ with $\beta=0.97,0.97,0.999$ for the experiments on CIFAR-10, respectively. As shown in the figures, the performance degrades very slightly with larger training intervals. It is noteworthy that even if we do not update the GP model after the warm-up phase (noted as $\Delta t=500$ for FMNIST, and $\Delta t=2000$ for CIFAR-10), FedCor still achieves faster convergence than the random selection strategy. These results indicate that the correlations learned by the GP model are stable, which supports our assumption in Section~\ref{SubSec:Train}. In a word, one can largely reduce the communication overhead by training the GP model with a very low frequency while guaranteeing the convergence rate and accuracy under the communication-bounded FL setting.

\subsection{Influence of Annealing Coefficient}
\begin{figure*}[tb]
\centering
\includegraphics[width=1.0\linewidth]{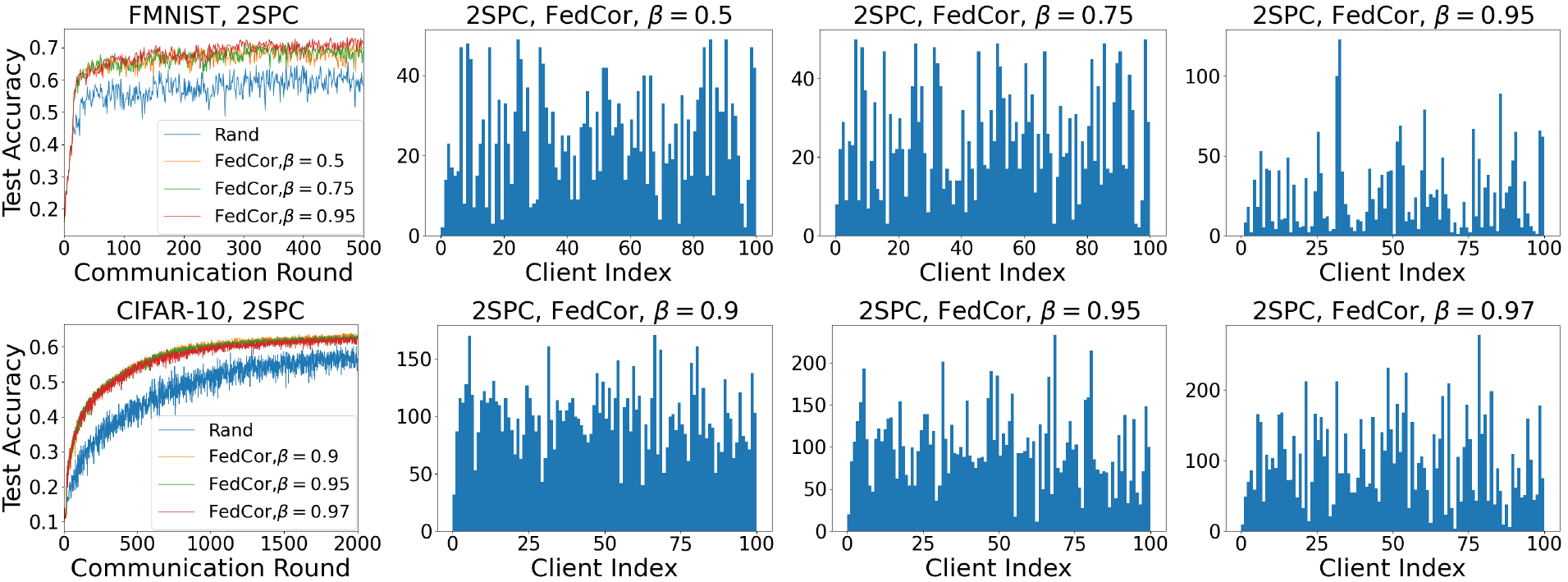}
\vspace{-1.5em}
\caption{Test accuracy and client selection frequency with different annealing coefficient $\beta$ on FMNIST and CIFAR-10 under the 2SPC setting. The frequency is represented as the number of times each client is selected during the whole training process.}
\label{fig:beta}
\end{figure*}
We also conduct experiments with different annealing coefficient $\beta$ that controls how ``concentrated'' the client selection is. We perform FedCor with $\Delta t = 10$ and $\beta = 0.5,0.75,0.9$ for FMNIST, and $\Delta t = 50, \beta = 0.9, 0.95, 0.99$ for CIFAR-10. The learning curves as well as the client selection frequencies under 2SPC setting are shown in \cref{fig:beta}, and we leave the full results under the 1SPC and Dir settings to \cref{Apx:beta}. We observe that when using a smaller $\beta$, the overall client selections appear to be more ``uniform'', while the learning curves are almost invariant. Notice that this does not mean that FedCor with small $\beta$ is equivalent to uniform sampling, instead, FedCor still achieves consistent improvements compared to uniform sampling. And \cref{SubSec:Insight} havs discussed the reason: FedCor not only considers the benefit that each client brings to the federation, but also considers the correlations among the clients to select the best group of clients. The experimental results here show that it is more important to select a good ``group'' of clients than just good individuals.

\subsection{Visualization of Client Embedding}
\begin{figure}[t]
  \centering
    \includegraphics[width=0.95\linewidth]{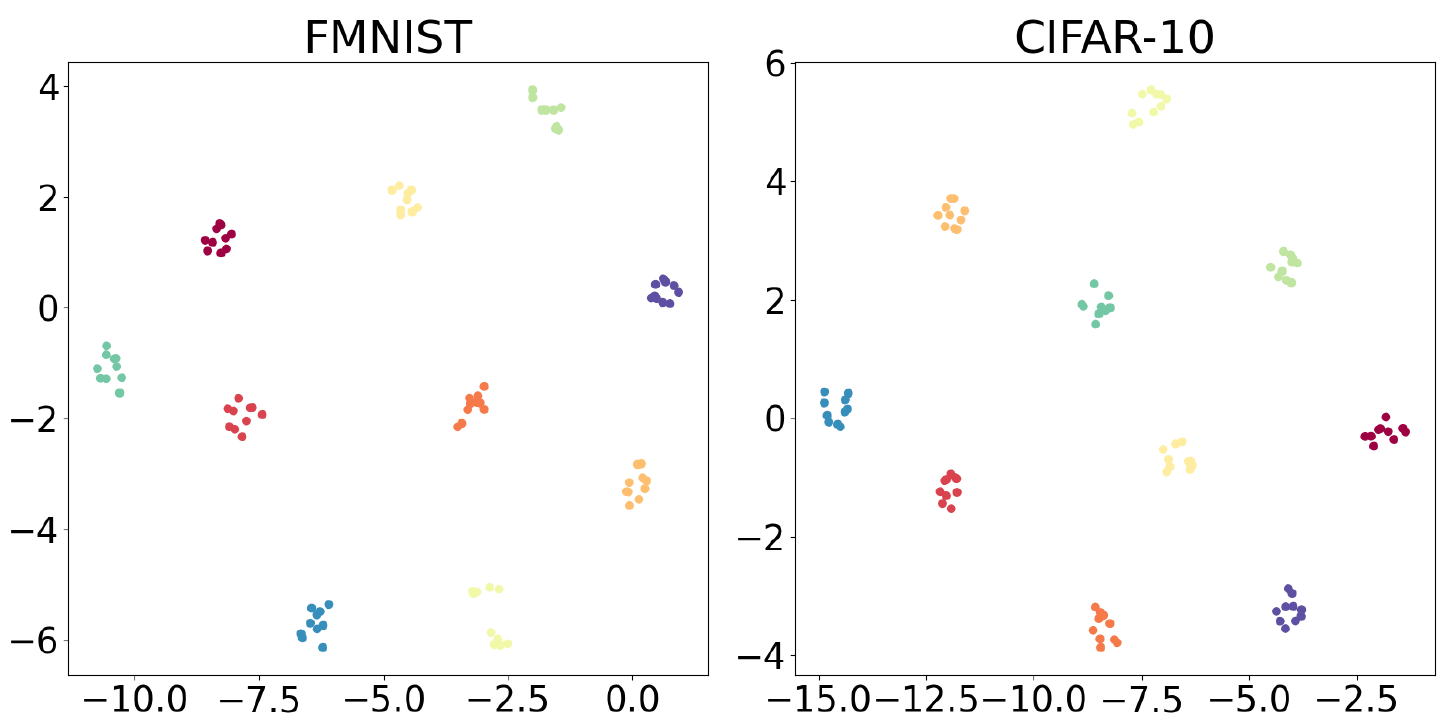}
  \caption{Visualization of client embedding under the 1SPC setting.
  }
\label{fig:visualization}
\end{figure}

To obtain an insight into the correlations learned by the GP model, we show the t-SNE~\cite{maaten2008visualizing} plot of the client embeddings learned in the warm-up phase under the 1SPC setting. In \cref{fig:visualization}, each embedding is labeled with the only data label on the corresponding client. We normalize the length of embedding vectors to $1$ so that the distance between two embeddings can reveal the correlation. We can see that the embeddings of clients with the same label are clustered together, which demonstrates that FedCor has captured the correlations between clients correctly in the warm-up phase.

\section{Conclusion and Future Work}\label{Sec:Con}
This work proposes FedCor, an FL framework with a novel client selection strategy for heterogeneous settings. FedCor is based on the intuition that it is crucial to utilize the correlations between clients to achieve a faster and more stable convergence in heterogeneous FL. Specifically, we model the client correlations with a GP, and design an effective and interpretable client selection strategy based on it. We also develop a efficient method to train the GP with a low communication overhead. Experimental results on FMNIST and CIFAR-10 show that FedCor effectively accelerates and stabilizes the training process under highly heterogeneous settings. In addition, we verify that FedCor captures the client correlation correctly using only the loss information. How to extend FedCor to the other tasks and further utilize the captured correlations is an interesting direction for future work. Besides, our method focuses on the cross-silo federated learning scenario~\cite{kairouz2019advances}, and how to extend it to the cross-device scenario is a meaningful topic.


\section*{Acknowledgement}\label{Sec:ack}

This research was generously supported in part by Gift from Amazon, etc. Any opinions, conclusions or recommendations expressed in this material are those of the authors and do not reflect the views of Amazon and its contractors.

\clearpage
{\small
\bibliography{ref}
\bibliographystyle{ieee_fullname}
}

\clearpage
\appendix
\section{Theoretical Analysis}
\subsection{Analytical Insight into Gaussian Processes}\label{Apx:gp_analysis}
In this section, we give a mathematical explanation about why the loss changes obey Gaussian Distributions. Our analysis based on the following assumption where we assume that the global weight update in one communication round follow a Gaussian Distribution under uniformly client selection.

\begin{assumption}\label{assumption1}
In any communication round $t$, if the client selection $\sK_t$ is a random variable sampled from a uniform distribution, the global model update $\Delta\rvw^t(\sK_t) = \rvw^{t+1}(\sK_t)-\vw^t$ follows Gaussian Distribution, i.e.,
\begin{equation}
\begin{aligned}
    &\sK_t\sim {\rm Uniform}\big(\{\sK\subseteq \sU:\vert\sK\vert=C\}\big)\\
    \Rightarrow&\Delta \rvw^t(\sK_t)\sim\mathcal{N}(\Delta \vw^t;-\eta_t \tilde{\vg}^t,\frac{\eta_t^2\mB\mB^T}{C}),
\end{aligned}
\end{equation}
where $\tilde{\vg}^t = \mathbb{E}_\rk[\tilde\nabla l_\rk(\vw^t)]$ is the mean cumulative gradient of all the clients in $\sU$, and $\mB$ is a constant matrix.
\end{assumption}


Assumption~\ref{assumption1} is inspired by \cite{mandt2016variational} who assumes the stochastic gradients in SGD are Gaussian, and therefore the parameter update after one iteration follows a Gaussian Distribution. Note that in the FL procedure, the form in Eq.~\ref{eq:fl_update} is very similar to that in the SGD update. The only difference is that the average gradients within one mini-batch is replaced by the average cumulative gradients of the selected clients. Therefore, it is reasonable to make this assumption similar to \cite{mandt2016variational}. 

To make a distinction, we use $\Delta \rvw$ without parentheses to denote a random variable w.r.t. the uniformly sampled client selection, and use $\Delta \vw(\sK)$ to denote a determinate value without randomness where the client selection $\sK$ is determined. The rule for $\Delta\rvl$ and $\Delta \vl(\sK)$ in the following contents is the same.

Based on this assumption, we can easily show that the loss changes in each communication round follow a Gaussian Process under first-order approximation, with the property of Gaussian Distribution.

\begin{corollary}\label{lemma1}
In any communication round $t$, $\forall\sS=\{i_1,\cdots,i_{\vert \sS\vert}\}\subseteq \sU$, the loss changes $\Delta \rvl_\sS^t = [\Delta \rl_{i_1}^t,\cdots,\Delta \rl_{i_{\vert \sS\vert}}^t]^T$ follow a Multivariate Gaussian Distribution (or a Gaussian Process) under first-order approximation, i.e.,
\begin{equation}
\begin{aligned}
&\Delta \rvl_{\sS}^t \sim \mathcal{N}(\Delta \vl^t_{\sS};\boldsymbol\mu_{\sS}^t,{\bm{\Sigma}}_{\sS}^t),\\
&\text{where }\\
&\bm{\mu}_{\sS}^t = -\eta_t{\mG_{\sS}^t}^T\tilde{\vg}^t;\\
&{\bm{\Sigma}}_{\sS}^t = \frac{\eta_t^2}{C}{\mG_{\sS}^t}^T\mB\mB^T\mG_{\sS}^t;\\
&\mG_{\sS}^t = \begin{bmatrix}\nabla l_{i_1}(\vw^t),\cdots,\nabla l_{i_{\vert {\sS}\vert}}(\vw^t)\end{bmatrix}.
\label{eq:GP_Lemma}
\end{aligned}
\end{equation}
\end{corollary}
We remove the subscript $\sS$ to simplify the corresponding representation for the client set $\sU$ as 
\begin{equation}
\Delta \rvl^t\sim\mathcal{N}(\Delta \vl^t;\bm{\mu}^t,{\bm{\Sigma}}^t),
\end{equation}
which is exactly the result in Eq.~\ref{eq:GP_all}. And we can also obtain a mathematical reason from Eq.~\ref{eq:GP_Lemma} for our choice of homogeneous linear kernel in Section~\ref{SubSec:Train}, where $\mX^t=\mB^T\mG^t$.

\paragraph{Remark} Although an uniformly sampled client selection is required in Assumption~\ref{assumption1} to get the loss changes to follow a GP prior, it is not necessary for the final selection to be uniformly sampled since we are predicting its loss changes with the GP posterior conditioned on the selected clients. We can view each posterior during the iterative selection process in Section~\ref{SubSec:Sel} as the distribution of the loss changes w.r.t. the client selection that consists of two parts: (i) fixed selected clients in the previous iteration and (ii) uniformly sampled clients from the rest of the clients.

\subsection{Proof of Lemma~\ref{lemma2}}\label{Apx:lemma2_proof}
To prove Lemma~\ref{lemma2}, we first introduce another assumption.
\begin{assumption}\label{assumption3}
In any communication round $t$, for any client selection $\sK$, we have
\begin{equation}
    Pr(\sK|\Delta \vl_{\sK}^t(\sK))\approx 1.\label{eq:condition_loss}
\end{equation}
\end{assumption}
This assumption asserts that for any client selection $\sK$, there is unlikely another client selection other than $\sK$ which can produce the same loss changes on $\sK$, i.e.,
\begin{equation}
\begin{aligned}
    &\forall \sK',\sK\subseteq \sU,\vert\sK'\vert=\vert\sK\vert\\
    \Rightarrow &Pr(\Delta \vl_{\sK}^t(\sK')= \Delta \vl_{\sK}^t(\sK)|\sK'\neq\sK)\approx 0.
\end{aligned}
\end{equation}
We anticipate that this is a realistic assumption because of the heterogeneity between clients and the highly complexity of the neural network. When selecting different clients, the data used for training varies a lot under heterogeneous federated learning settings. This fact makes it almost impossible to produce the same neural network, and thus the same loss changes, with two different client selections. Furthermore, the selected clients usually have larger loss decreases than other clients who are not selected, because the model update is based on the mean cumulative gradient of these selected clients. The other client selection is unlikely to generate the same large loss decreases on all of them.

With Assumption~\ref{assumption3}, we can get the following corollary~\ref{corollary1}.
\begin{corollary}\label{corollary1}
In any communication round $t$, for any client selection $\sK$, we have
\begin{equation}
    Pr(\Delta \vl^t(\sK)|\Delta \vl_{\sK}^t(\sK))\approx 1.\label{eq:loss_condition_loss}
\end{equation}

\begin{proof}
When client selection $\sK$ is given, we get the determinate model update $\Delta\vw^t(\sK)$, thus the loss changes are known without randomness. In the other word,
\begin{equation}
    Pr(\Delta \vl^t(\sK)|\sK)=1\label{eq:condition_selection}
\end{equation}
always holds. Besides, we can extend the condition in Eq.~\ref{eq:condition_loss} to the loss changes of all the clients and get
\begin{equation}
    Pr(\sK|\Delta \vl^t(\sK))\approx 1.\label{eq:condition_loss_all}
\end{equation}
Combining Eq.~\ref{eq:condition_loss}, Eq.~\ref{eq:condition_selection} and Eq.~\ref{eq:condition_loss_all}, we have
\begin{align}
     Pr(\Delta \vl^t(\sK))\approx &Pr(\Delta \vl^t(\sK),\sK)\\
     =&Pr(\sK)\\
     =& Pr(\Delta \vl_{\sK}^t(\sK),\sK)\\
     \approx &Pr(\Delta \vl_{\sK}^t(\sK))\label{eq:condition_eq}
\end{align}
By substituting Eq.~\ref{eq:condition_eq} into the expression of $Pr(\Delta \vl^t(\sK)|\Delta \vl_{\sK}^t(\sK))$, we get
\begin{align}
    Pr(\Delta \vl^t(\sK)|\Delta \vl_{\sK}^t(\sK))
    =&\frac{Pr(\Delta \vl^t(\sK),\Delta \vl_{\sK}^t(\sK))}{Pr(\Delta \vl_{\sK}^t(\sK))}\\
    =&\frac{Pr(\Delta \vl^t(\sK))}{Pr(\Delta \vl_{\sK}^t(\sK))}\\
    \approx &1.
\end{align}
\end{proof}
\end{corollary}

Now we are ready to prove Lemma~\ref{lemma2}.
\setcounter{lemma}{0}
\begin{lemma}
The optimization problem in \cref{eq:opt} is approximately equivalent to the following probabilistic form.
\begin{equation}\label{eq:gp_objective_pf}
    \min_{\sK_{t}}\quad\mathbb{E}_{\Delta \rvl^t|\Delta\bm l^{t}_{\sK_{t}}(\sK_t)}\Big[\sum_i p_i\Delta l^{t}_i\Big] = 
    \sum_i p_i\tilde{\mu}^{t}_i(\Delta  \bm l^{t}_{\sK_{t}}(\sK_t)),
\end{equation}
where $\Delta \rvl^t=[\Delta\rl^t_1,\cdots,\Delta\rl^t_N]$ is the loss changes of all clients in round $t$, which is a random variable w.r.t random client selection in round $t$. $\tilde{\bm{\mu}}^{t}(\Delta  \bm l^{t}_{\sK_{t}}(\sK_t))$ is the posterior mean of $\Delta\rvl^t$ conditioned on $\Delta  \bm l^{t}_{\sK_{t}}(\sK_t)=[\Delta l^t_i(\sK_t)]_{i\in \sK_t}$.

\begin{proof}
According to Corollary~\ref{corollary1}, we can transform the optimization problem in Eq.~\ref{eq:opt} into the form in Eq.~\ref{eq:gp_objective_pf}.
\begin{align}
    &\min_{\sK_t}\quad \Delta L^{t}(\sK_t)\\
    =&\min_{\sK_t}\quad \sum_{i}p_i\Delta l_i^t(\sK_t)\\
    \approx &\min_{\sK_t}\quad Pr(\Delta \vl^t(\sK_t)|\Delta \vl_{\sK_t}^t(\sK_t))\sum_{i}p_i\Delta l_i^t(\sK_t)\\
    \approx &\min_{\sK_t}\quad \mathbb{E}_{\Delta \rvl^t|\Delta\vl^{t}_{\sK_{t}}(\sK_t)}\Big[\sum_i p_i\Delta l^{t}_i\Big]\\
    =&\min_{\sK_t}\quad \sum_i p_i\tilde{\mu}^{t}_i(\Delta  \vl^{t}_{\sK_{t}}(\sK_t)).
\end{align}
\end{proof}
\end{lemma}

\subsection{Proof of Lemma~\ref{lemma3}}\label{Apx:lemma3_proof}
\begin{lemma}
The selection criterion of FedCor when selecting two clients $k_1$ and $k_2$ can be written as
\begin{align}
    k_1&=\mathop{\arg\max}_k\quad \beta^{\tau_k}\sum_ip_i\sigma_ir_{ik},\label{eq:sim_selection_pf}\\
    k_2&=\mathop{\arg\max}_{k'}\quad \frac{\beta^{\tau_{k'}}\Big[\sum_ip_i \sigma_{i}r_{ik'}-r_{k_1k'}\sum_ip_i \sigma_ir_{ik_1}\Big]}{\sqrt{1-r_{k'k_1}^2}},\label{eq:post_criterion_pf}
\end{align}
where $r_{ij}=\Sigma_{i,j}/\sigma_i\sigma_j$ is the Pearson correlation coefficient.

\begin{proof}
We first deduce Eq.~\ref{eq:sim_selection_pf} for the first client $k_1$. By substituting the loss change estimation $\Delta \hat l_k$ from Eq.~\ref{eq:guess} into the criterion in Eq.~\ref{eq:selection}, we can calculate the weighted sum of the posterior mean as
\begin{align}
&\sum_ip_i\tilde{\mu}_i(\Delta \hat l_k)\\ 
= &\sum_ip_i\mu_i+\sum_ip_i\frac{\Sigma_{i,k}}{\sigma_k^2}(\Delta \hat l_k-\mu_k)\\
=& \sum_ip_i\mu_i-a\beta^{\tau_k}\sum_ip_i\sigma_ir_{ik},\label{eq:extend_selection}
\end{align}
where $r_{ik}$ is the Pearson correlation coefficient. The first item in Eq.~\ref{eq:extend_selection} and the factor $a$ are constant for all $k$, thus the selection strategy becomes
\begin{equation}\label{eq:sim_selection_pf_mid}
    k_1=\mathop{\arg\max}_k\quad \beta^{\tau_k}\sum_ip_i\sigma_ir_{ik},
\end{equation}
which is Eq.~\ref{eq:sim_selection_pf}.

Then we deduce Eq.~\ref{eq:post_criterion_pf} for selecting $k_2$. We can calculate the posterior covariance conditioned on $\Delta\hat l_{k_1}$ as
\begin{align}
    \tilde{\Sigma}_{i,j}(\Delta\hat l_{k_1})&=\Sigma_{i,j}-\frac{\Sigma_{i,k_1}\Sigma_{k_1,j}}{\sigma_{k_1}^2}\\
    &=\sigma_i\sigma_j(r_{ij}-r_{ik_1}r_{k_1j})\label{eq:post_cov}\\
    \tilde{\sigma}_i(\Delta\hat l_{k_1}) &= \sqrt{\tilde{\Sigma}_{i,i}(\Delta\hat l_{k_1})}\\
    &=\sigma_i\sqrt{1-r_{ik_1}^2}\label{eq:post_std}.
\end{align}
We substitute the posterior covariance into the simplified selection criterion in Eq.~\ref{eq:sim_selection_pf_mid} and get
\begin{equation}
\begin{aligned}
&\beta^{\tau_{k'}}\sum_ip_i\frac{\tilde{\Sigma}_{i,k'}(\Delta\hat l_{k_1})}{\tilde{\sigma}_{k'}(\Delta\hat l_{k_1})}\\
=&\frac{\beta^{\tau_{k'}}\Big[\sum_ip_i\sigma_{i} r_{ik'}-r_{k_1k'}\sum_ip_i \sigma_ir_{ik_1}\Big]}{\sqrt{1-r_{k'k_1}^2}}.
\end{aligned}
\end{equation}
So we have Eq.~\ref{eq:post_criterion_pf}:
\begin{equation}
k_2=\mathop{\arg\max}_{k'}\quad \frac{\beta^{\tau_{k'}}\Big[\sum_ip_i\sigma_{i} r_{ik'}-r_{k_1k'}\sum_ip_i \sigma_ir_{ik_1}\Big]}{\sqrt{1-r_{k'k_1}^2}}.
\end{equation}
\end{proof}

\end{lemma}

\section{Selection Criterion and Convergence Analysis}\label{Apx:convergence}
In this section, we will analyse FedCor when selecting arbitrary number of clients. While the iterative client selection makes it obscure to analyse the convergence, we will show that we can construct a simpler proxy algorithm who can approximate the selection strategy of FedCor and there for share similar convergence characteristic. We will prove the convergence of this proxy algorithm.

\subsection{Definitions}
We first introduce some important definitions. In the following analysis, We denote the client selection sampled from FedCor as $\sK_t\sim \pi$ and client selection sampled uniformly as $\sK_t\sim \mathcal{U}$.

In the $j$-th iteration of FedCor, we select a client $k_j$ to minimize the posterior mean of the loss change. Since the prior mean in each iteration is fixed, we can say that we are maximizing the decrease from prior mean $\bm{\mu}^{t,j}$ to posterior mean $\tilde{\bm{\mu}}^{t,j}$. We define the posterior gain of this iteration as the decrease from prior mean to posterior mean, namely,
\begin{align}
    g^{t,j}(k_j)&=\sum_ip_i(\mu^{t,j}_i-\tilde{\mu}^{t,j}_i(\Delta\hat {l}^t_{k_j}))\\
    &=\alpha_{k_j}^{t}\sum_ip_i\sigma_i^{t,j}r_{ik_{j}}^{t,j}.
\end{align}
We define $\bm{\mu}^{t,1}=\bm{\mu}^t$ and $\bm{\Sigma}^{t,1}=\bm{\Sigma}^t$. And for $j>1$ we have 
\begin{align}
    \bm{\mu}^{t,j}=\tilde{\bm{\mu}}^{t,j-1}(\Delta\hat {l}^t_{k_{j-1}}),\qquad
    \bm{\Sigma}^{t,j}=\tilde{\bm{\Sigma}}^{t,j-1}(\Delta\hat {l}^t_{k_{j-1}}).
\end{align}
With Lemma~\ref{lemma3}, we get
\begin{align}
    g^{t,j}(k_j)=\frac{g^{t,j-1}(k_j)-\frac{\alpha_{k_j}^t}{\alpha_{k_{j-1}}^t}r^{t,j-1}_{k_{j-1}k_j}g^{t,j-1}(k_{j-1})}{\sqrt{1-{r^{t,j-1}_{k_{j-1}k_j}}^2}}.\label{eq:gain}
\end{align}
With this notation, we can simplify our selection strategy as follows.
\begin{align}
    k_j^*=\mathop{\arg\max}_{k_j}g^{t,j}(k_j).
\end{align}

We further define the one-round advantage of FedCor compared with uniform sampling as follows.
\begin{align}
    A^t=&\mathbb{E}_{\sK_t\sim \mathcal{U}}[L(\vw^{t+1})-L(\vw^t)]-\nonumber\\
    &\mathbb{E}_{\sK_t\sim\pi}[L(\vw^{t+1})-L(\vw^{t})]\\
    =&\sum_{j=1}^C g^{t,j}(k_{j}^*)\label{eq:adv}.
\end{align}
The second equation directly arises from the definition of our prior distribution where $\mathbb{E}_{\sK_t\sim \mathcal{U}}[L(\vw^{t+1})-L(\vw^t)]=\sum_i\mu_i^{t}$. 

Unfortunately, because of the iterative selection, the selection criterion of $k_j$ depends on the previous selected clients, which makes a quantitatively analysis complicated. To bypass this difficulty, we will first point out that $A^t$ has a lower bound that is tight in some special cases. We find that a proxy client selection strategy that maximizes this lower bound has a similar but simpler behaviour compared with FedCor, and we will also give a convergence guarantee of the proxy algorithm. 

\subsection{Approximation of FedCor}

An important property of FedCor is that it prefers clients who have lower correlations with those selected in the previous iteration, since
\begin{equation}
    \forall r^{t,j-1}_{k_{j-1}k_j}\in (-1,1), \frac{\partial g^{t,j}(k_j)}{\partial r^{t,j-1}_{k_{j-1}k_j}}<0.
\end{equation}
We further predict that FedCor tends to select clients that with $r^{t,j-1}_{k_{j-1}k_j}$ close to $0$ instead of $r^{t,j-1}_{k_{j-1}k_j}<0$ because if $r^{t,j-1}_{k_{j-1}k_j}<0$, $k_j$ should be far away from $k_{j-1}$ who is closed to other clients in the embedding space, which makes $k_j$ has low correlation with the other clients and not be selected. Therefore, we can infer that FedCor will select a group of clients who have nearly zero correlations with each other, which simplifies the expression of $g^{t,j}(k_j)$ to $g^{t,1}(k_j)$. 

Based on the analysis above, we define a proxy algorithm $\tilde \pi$ who maximize the following objective.
\begin{align}
    \tilde A^t=\sum_{k_j\in\sK_t}^C g^{t,1}(k_j)\approx\sum_{k\in\sK_t}\sum_ip_i\Sigma^t_{i,k},
\end{align}
where we further omit the difference of $\alpha_{k}^t$ and $\sigma_{k}^t$ for different client $k$. We can use the client selection generated by this proxy algorithm to approximate the client selection of FedCor, and thus they share similar convergence characteristic.

In the following section, we will show that this proxy algorithm has a good property that enable it to converge to the optimal solution of the global loss $L$ without gap, even it is a biased selection strategy.



\subsection{Convergence Analysis of the Proxy Algorithm}
In the following section, we denote the client selection sampled from the proxy client selection strategy as $\sK_t\sim\tilde\pi$. We use $\mathbb{E}[\cdot]$ as the expectation over the mini-batch and $\mathbb{E}_{\sK_t}[\cdot]$ as the expectation over the client selection strategy. We first give the common assumptions used in Federated Learning~\cite{li2019convergence,cho2020client}.

\begin{assumption}\label{assumption_c1}
$l_1,l_2,\cdots,l_N$ are all $M$-smooth: for all $\vv$ and $\vw$, $l_k(\vv)\le l_k(\vw)+(\vv-\vw)^T\nabla l_k(\vw)+\frac{M}{2}\Arrowvert \vv-\vw\Arrowvert_2^2$. 
\end{assumption}

\begin{assumption}\label{assumption_c2}
$l_1,l_2,\cdots,l_N$ are all $m$-strongly convex: for all $\vv$ and $\vw$, $l_k(\vv)\ge l_k(\vw)+(\vv-\vw)^T\nabla l_k(\vw)+\frac{m}{2}\Arrowvert \vv-\vw\Arrowvert_2^2$. 
\end{assumption}

\begin{assumption}\label{assumption_c3}
For the mini-batch $\xi_k\in\sD_k$ sampled uniformly on each client $k\in\sU$, the variance of stochastic gradients is bounded: $\mathbb{E}\Arrowvert\nabla l_k(\vw_k,\xi_k)-\nabla l_k(\vw_k)\Arrowvert^2\le s_k^2$.
\end{assumption}

\begin{assumption}\label{assumption_c4}
For each client $k\in\sU$ and any communication round $t$, the expected squared norm of stochastic gradients is uniformly bounded: $\mathbb{E}\Arrowvert\nabla l_k(\vw_k,\xi_k)\Arrowvert^2\le G^2$.
\end{assumption}

For concision, we omit $\mathbb{E}$ in the following content and apply an expectation over the mini-batch by default.

Now we give an important property of the proxy algorithm that will be used for proving the convergence.
\begin{lemma}\label{lemma6}
In any communication round $t$, with Assumption~\ref{assumption1} and Assumption~\ref{assumption3} holds, we have
\begin{align}\label{eq:theorem1}
    \sK_t\sim\tilde\pi=\mathop{\arg\max}_{\sK}(\mB^T\nabla L(\vw^t))^T\sum_{k\in\sK}\mB^T\nabla l_k(\vw^t).
\end{align}
\begin{proof}
In the proxy algorithm, we have
\begin{align}
    \sK_t=&\mathop{\arg\max}_{\sK}\sum_{k\in\sK_t}\sum_ip_i\Sigma^t_{i,k}\\
    =&\mathop{\arg\max}_{\sK}\frac{\eta^2_t}{C}\sum_{k\in\sK}\sum_i p_i\nabla l_i(\vw^t)\mB\mB^T\nabla l_k(\vw^t)\label{eq:theorem1_pf_1}\\
    =&\mathop{\arg\max}_{\sK}(\mB^T\nabla L(\vw^t))^T\sum_{k\in\sK}\mB^T\nabla l_k(\vw^t).\label{eq:theorem1_pf_2}
\end{align}
Eq.~\ref{eq:theorem1_pf_1} comes from the expression of $\bm{\Sigma}^t$ in \cref{lemma1}, and \cref{eq:theorem1_pf_2} arises from $L(\vw^t)=\sum_i p_il_i(\vw^t)$. 
\end{proof}
\end{lemma}

To connect this property with the convergence of the algorithm, we first define a sequence and show that the convergence of this sequence is equivalent to the convergence of the algorithm with this property. We define Sequence $\Delta_t$ as follows.
\begin{align}
    &\Delta_t=\mathbb{E}_{\sK_t\sim\tilde\pi}\Arrowvert \vw^{t}-\vw_{\sK_{t}}^*\Arrowvert^2,\\
    &\text{where}\nonumber\\
    &\vw_{\sK_t}^*=\mathop{\arg\min}_{\vw}\sum_{k\in\sK_t}l_k(\vw).
\end{align}
We now show that if $\Delta_t\rightarrow 0$, we have $\vw\rightarrow \vw^*$.
\begin{corollary}\label{corollary3}
\textbf{(Optimal Solution Consistency)} If $\Delta_t$ converges to $0$, there must be $\vw^t$ converges to $\vw^*$.
\begin{align}
    \lim_{t\rightarrow\infty}\Delta_t=0\Rightarrow \lim_{t\rightarrow\infty}\vw^t=\vw^*
\end{align}
\begin{proof}
With $\sK_t\sim\tilde\pi$, we have
\begin{align}
    &\lim_{t\rightarrow\infty}\Delta_t=0\\
    \Rightarrow&\lim_{t\rightarrow\infty} \vw^t=\vw_{\sK_t}^*\\
    \Rightarrow&\lim_{t\rightarrow\infty} \sum_{k\in\sK_t}\nabla l_k(\vw^t)=\bm{0}\\
    \Rightarrow&\lim_{t\rightarrow\infty} (\mB^T\nabla L(\vw^t))^T\sum_{k\in\sK_t}\mB^T\nabla l_k(\vw^t)=0.
\end{align}
Since
\begin{align}
    \sK_t=\mathop{\arg\max}_{\sK}(\mB^T\nabla L(\vw^t))^T\sum_{k\in\sK}\mB^T\nabla l_k(\vw^t),
\end{align}
If $\lim_{t\rightarrow\infty}\mB^T\nabla L(\vw^t)\neq \bm{0}$ or does not converge, we can say that
\begin{align}
    &\forall \epsilon>0, \exists \tau, \forall t>\tau,\forall \sK,\\
    &(\mB^T\nabla L(\vw^t))^T\sum_{k\in\sK}\mB^T\nabla l_k(\vw^t)\le \epsilon,
\end{align}
which cannot be true since
\begin{align}
    \mathbb{E}_{\sK\sim\mathcal{U}}\sum_{k\in\sK}\nabla l_k(\vw^t)=C\nabla L(\vw^t).
\end{align}
Thus we conclude that 
\begin{align}
    \lim_{t\rightarrow\infty}\mB^T\nabla L(\vw^t)= 0.
\end{align}
If the Gaussian Distribution in Assumption~\ref{assumption1} is non-degenerate, we have
\begin{align}
    \lim_{t\rightarrow\infty}\nabla L(\vw^t)= \bm{0}\Rightarrow\lim_{t\rightarrow\infty}\vw^t=\vw^*
\end{align}
\end{proof}
\end{corollary}

We now only need to prove the convergence of $\Delta_t$, which will imply the convergence of the proxy algorithm according to Corollary~\ref{corollary3}. We first introduce one extra assumption as well as two lemmas that will be used in the proof.

For convenient, we define $L_{\sK_{t}}(\vw)=\frac{1}{C}\sum_{k\in\sK_t}l_k(\vw)$, and thus $\vw_{\sK_t}^*=\mathop{\arg\min}_{\vw}L_{\sK_{t}}(\vw)$. Notice that $\sK_t\sim\tilde\pi$ only depends on $\bm{\Sigma}^t$, thus we can say that $\vw_{\sK_t}^*$ is given by a function of $\bm{\Sigma}^t$, i.e., $\vw_{\sK_t}^*=\Omega(\bm{\Sigma}^t)$. We further assume the smoothness of $\Omega$:

\begin{assumption}\label{assumption_c5}
For any $t$, $\mathbb{E}\Arrowvert\vw_{\sK_{t+1}}^*-\vw_{\sK_{t}}^*\Arrowvert^2=\mathbb{E}\Arrowvert\Omega(\bm{\Sigma}^{t+1})-\Omega(\bm{\Sigma}^{t})\Arrowvert^2\le \delta\mathbb{E}\Arrowvert\bm{\Sigma}^{t+1}-\bm{\Sigma}^t\Arrowvert_1$, where $\Arrowvert\cdot\Arrowvert_1$ is the $\ell_1$ norm of a vector.
\end{assumption}

Now we introduce a lemma that bounds $\mathbb{E}\Arrowvert\bm{\Sigma}^{t+1}-\bm{\Sigma}^t\Arrowvert_1$.

\begin{lemma}\label{lemma7}
Assume \cref{assumption1}, \cref{assumption_c1} and \cref{assumption_c4}, if $\mathbb{E}\Arrowvert \vw^t-\vw^{t+1}\Arrowvert^2\le q_t^2$, we have
\begin{align}
    \mathbb{E}\Arrowvert \bm{\Sigma}_{t+1}-\bm{\Sigma}_{t}\Arrowvert_1 \le \frac{bN^2}{C}[\eta_t^2(G+Mq_t)^2-\eta_{t+1}^2G^2],
\end{align}
where $b$ is the largest eigenvalue of $\mB\mB^T$.
\begin{proof}
According to \cref{assumption1}, we have
\begin{align}
    \Sigma_{i,j}=\frac{\eta_t^2}{C}\nabla {l_i^t}^T \mB\mB^T \nabla l_j^t.
\end{align}
And we can calculate
\begin{align}
    &\left|\Sigma_{i,j}^{t+1}-\Sigma_{i,j}^t\right|\\
    =&\Big|\frac{\eta_t^2}{C}(\nabla {l_i^{t+1}}^T\mB\mB^T\nabla l_j^{t+1}-\nabla {l_i^{t}}^T\mB\mB^T\nabla l_j^{t})+\nonumber\\
    &\frac{\eta_{t+1}^2-\eta_t^2}{C}\nabla {l_i^{t+1}}^T\mB\mB^T\nabla l_j^{t+1}\Big|\\
    \le&\frac{\eta_t^2}{C}\left|\nabla {l_i^{t+1}}^T\mB\mB^T\nabla l_j^{t+1}-\nabla {l_i^{t}}^T\mB\mB^T\nabla l_j^{t}\right|+\nonumber\\
    &\frac{\eta_{t}^2-\eta_{t+1}^2}{C}\left|\nabla {l_i^{t+1}}^T\mB\mB^T\nabla l_j^{t+1}\right|.\label{eq:lemma7_eq1_pf}
\end{align}
We now bound each term in \cref{eq:lemma7_eq1_pf} separately. For the first term,
\begin{align}
    &\left|\nabla {l_i^{t+1}}^T\mB\mB^T\nabla l_j^{t+1}-\nabla {l_i^{t}}^T\mB\mB^T\nabla l_j^{t}\right|\\
    =&\Big|(\nabla {l_i^{t+1}}-\nabla {l_i^{t}})^T\mB\mB^T\nabla l_j^{t}+\nonumber\\
    &\nabla {l_i^{t}}^T\mB\mB^T(\nabla {l_j^{t+1}}-\nabla {l_j^{t}})+\nonumber\\
    &(\nabla {l_i^{t+1}}-\nabla {l_i^{t}})^T\mB\mB^T(\nabla {l_j^{t+1}}-\nabla {l_j^{t}})\Big|\\
    \le&b\Big(\Arrowvert\nabla {l_i^{t+1}}-\nabla {l_i^{t}}\Arrowvert\Arrowvert\nabla l_j^{t}\Arrowvert+\nonumber\\
    &\Arrowvert\nabla {l_j^{t+1}}-\nabla {l_j^{t}}\Arrowvert\Arrowvert\nabla l_i^{t}\Arrowvert+\nonumber\\
    &\Arrowvert\nabla {l_i^{t+1}}-\nabla {l_i^{t}}\Arrowvert\Arrowvert\nabla {l_j^{t+1}}-\nabla {l_j^{t}}\Arrowvert\Big)\\
    \le&b\Big[M\Arrowvert\vw^{t+1}-\vw^{t}\Arrowvert(\Arrowvert\nabla l_j^{t}\Arrowvert+\Arrowvert\nabla l_i^{t}\Arrowvert)+\nonumber\\
    &M^2\Arrowvert \vw^{t+1}-\vw^{t}\Arrowvert^2\Big],
\end{align}
where $b$ is the largest eigenvalue of $\mB\mB^T$. For the second term,
\begin{align}
    \left|\nabla {l_i^{t+1}}^T\mB\mB^T\nabla l_j^{t+1}\right|\le b\Arrowvert\nabla {l_i^{t+1}}\Arrowvert\Arrowvert\nabla {l_j^{t+1}}\Arrowvert.
\end{align}
We take the expectation over both sides and with Cauchy-Schwarz inequality, we get
\begin{align}
    &\mathbb{E}\left|\Sigma_{i,j}^{t+1}-\Sigma_{i,j}^t\right|\\
    \le& \frac{\eta_t^2}{C}b\Big[M\sqrt{\mathbb{E}\Arrowvert \vw^{t+1}-\vw^{t}\Arrowvert^2\mathbb{E}\Arrowvert\nabla l_j^{t}\Arrowvert^2}+\nonumber\\
    &M\sqrt{\mathbb{E}\Arrowvert \vw^{t+1}-\vw^{t}\Arrowvert^2\mathbb{E}\Arrowvert\nabla l_i^{t}\Arrowvert^2}+\nonumber\\
    &M^2\mathbb{E}\Arrowvert \vw^{t+1}-\vw^{t}\Arrowvert^2\Big]+\nonumber\\
    &\frac{\eta_{t}^2-\eta_{t+1}^2}{C}b\sqrt{\mathbb{E}\Arrowvert\nabla {l_i^{t+1}}\Arrowvert^2\mathbb{E}\Arrowvert\nabla {l_j^{t+1}}\Arrowvert^2}\\
    \le& \frac{\eta_t^2b}{C}(G^2+2Mq_tG+M^2q_t^2)-\frac{\eta_{t+1}^2}{C}bG^2
\end{align}
And we have
\begin{align}
    &\mathbb{E}\Arrowvert \bm{\Sigma}_{t+1}-\bm{\Sigma}_{t}\Arrowvert_1\\
    =&\sum_{i,j}^N \mathbb{E}\left|\Sigma_{i,j}^{t+1}-\Sigma_{i,j}^t\right|\\
    \le& \frac{bN^2}{C}[\eta_t^2(G+Mq_t)^2-\eta_{t+1}^2G^2]
\end{align}
\end{proof}
\end{lemma}

We will also use the following lemma that is proved by \cite{li2019convergence}.
\begin{lemma}\label{lemma5}
Assume Assumption~\ref{assumption_c1} to \ref{assumption_c4}. If $\eta_t\le \frac{1}{4M}$, with full and balanced participation in FedAvg, in any communication round $t$ and its $i$-th iteration, we have
\begin{align}
    \mathbb{E}\Arrowvert \bar\vw^{t,i+1}-\vw^*\Arrowvert^2\le(1-\eta_tm)\mathbb{E}\Arrowvert \bar\vw^{t,i}-\vw^*\Arrowvert^2+\eta_t^2F,
\end{align}
where
\begin{align}
    &F=\frac{1}{N}\sum_{k=1}^N s_k^2+6M\Gamma+8(E-1)^2G^2,\\
    &\Gamma = L^*-\frac{1}{N}\sum_{k=1}^{N}l_k^*.
\end{align}
Here, $\bar\vw^{t,i}=\frac{1}{N}\sum_{k=1}^N \vw_k^{t,i}$, and $\vw^{t,i}_k$ is the local weight at the $i$-th iteration of communication round $t$. $E$ is the total number of local training iterations. $L^*=L(\omega^*)$ and $l_k^*=l_k(\omega_k^*)$ are the optimal value of $L$ and $l_k$, respectively.
\end{lemma}

Now we give the theorem of the convergence of $\Delta_t$ and prove it.
\begin{theorem}
With Assumption~\ref{assumption1} to \ref{assumption_c5} holds, with learning rate $\eta_t=\frac{\beta}{t+\gamma}$ for some $\beta>\frac{1}{m}$ and $\gamma>0$ such that $\eta_1\le \min\{\frac{1}{m},\frac{1}{4M}\}=\frac{1}{4M}$, we have
\begin{align}
    \Delta_t\le\frac{\nu}{\gamma+t},
\end{align}
where 
\begin{align}
    &\nu = \max\{\frac{\beta^2(\tilde F+\tilde D)}{\beta m-1},(\gamma+1)\Delta_1\},\\
    &\tilde F=2E\max_{t}F_t,\\
    &F_t=\frac{1}{C}\sum_{k\in\sK_t} s_k^2+6M\Gamma_t+8(E-1)^2G^2,\\
    &\Gamma_t=L_{\sK_t}^*-\frac{1}{C}\sum_{k\in\sK_t}l_k^*,\\
    &\tilde D=(\frac{1}{m}+\frac{1}{4M})\delta D,\\
    &D=\frac{bN^2}{C}(2mG^2+2MEG+\frac{1}{4}ME^2G^2).
\end{align}

\begin{proof}
For $\sK_t\sim\tilde\pi(\vw^{t})$ and $\sK_{t+1}\sim\tilde\pi(\vw^{t+1})$, we have 
\begin{align}
    \Delta_{t+1}=&\Arrowvert \vw^{t+1}-\vw_{\sK_{t+1}}^*\Arrowvert^2\\
    =& \Arrowvert \vw^{t+1}-\vw_{\sK_{t}}^*\Arrowvert^2+\Arrowvert \vw_{\sK_{t}}^*-\vw_{\sK_{t+1}}^*\Arrowvert^2+\nonumber\\
    &2\langle\vw^{t+1}-\vw_{\sK_{t}}^*,\vw_{\sK_{t}}^*-\vw_{\sK_{t+1}}^*\rangle\\
    \le& \Arrowvert \vw^{t+1}-\vw_{\sK_{t}}^*\Arrowvert^2+\Arrowvert \vw_{\sK_{t}}^*-\vw_{\sK_{t+1}}^*\Arrowvert^2+\nonumber\\
    &\eta_t m\Arrowvert \vw^{t+1}-\vw_{\sK_{t}}^*\Arrowvert^2+\nonumber\\
    &\frac{1}{\eta_tm}\Arrowvert \vw_{\sK_{t}}^*-\vw_{\sK_{t+1}}^*\Arrowvert^2\label{eq:convergence_pf_0}\\
    \le&(1+\eta_tm)\Arrowvert \vw^{t+1}-\vw_{\sK_{t}}^*\Arrowvert^2+\nonumber\\
    &(1+\frac{1}{\eta_tm})\delta\Arrowvert\bm{\Sigma}^{t+1}-\bm{\Sigma}^t\Arrowvert_1,\label{eq:convergence_pf_1}
\end{align}
where Eq.~\ref{eq:convergence_pf_0} arises from AM-GM inequality and Eq.~\ref{eq:convergence_pf_1} arises from Assumption~\ref{assumption_c5}.

For the first term in Eq.~\ref{eq:convergence_pf_1}, we can bound it by Lemma~\ref{lemma5} as follows. The key point here is that when training in one communication round $t$, we can view this round a small FL process with clients in $\sK_t$ fully participating. In this view, the global loss and the optimal global weight becomes $L_{\sK_t}$ and $\vw_{\sK_t}^*$ instead. Thus we can apply Lemma~\ref{lemma5} directly to bound $\Arrowvert \vw_{t+1}-\vw_{\sK_{t}}^*\Arrowvert^2$. With $\eta_t\le \frac{1}{4M}\le\frac{1}{m}$, we have $\eta_tm\le1$ and $1+\eta_tm\le\frac{1}{1-\eta_tm}$, and we can get

\begin{align}
    &(1+\eta_tm)\Arrowvert \vw^{t+1}-\vw_{\sK_{t}}^*\Arrowvert^2\\
    =&(1+\eta_tm)[\Arrowvert \bar\vw^{t,E}-\vw_{\sK_{t}}^*\Arrowvert^2]\\
    \le&(1+\eta_tm)[(1-\eta_tm)\Arrowvert \bar\vw^{t,E-1}-\vw_{\sK_{t}}^*\Arrowvert^2+\eta_t^2F_t]\\
    \le&(1+\eta_tm)\{(1-\eta_tm)^2\Arrowvert \bar\vw^{t,E-2}-\vw_{\sK_{t}}^*\Arrowvert^2+\nonumber\\
    &[1+(1-\eta_tm)]\eta_t^2F_t\}\\
    &\cdots\nonumber\\
    \le&(1+\eta_tm)\Big\{(1-\eta_tm)^E\Arrowvert \bar\vw^{t,0}-\vw_{\sK_{t}}^*\Arrowvert^2+\nonumber\\&[1+(1-\eta_tm)+\cdots+(1-\eta_tm)^{E-1}]\eta_t^2F_t\Big\}\\
    \le&(1-\eta_tm)^{E-1}\Arrowvert \vw^{t}-\vw_{\sK_{t}}^*\Arrowvert^2+\nonumber\\
    &(1+\eta_tm)\frac{1-(1-\eta_tm)^E}{m}\eta_tF_t\\
    \le& (1-\eta_tm)\Arrowvert \vw^{t}-\vw_{\sK_{t}}^*\Arrowvert^2+(1+\eta_tm)E\eta_t^2F_t\\
    \le& (1-\eta_tm)\Arrowvert \vw^{t}-\vw_{\sK_{t}}^*\Arrowvert^2+2E\eta_t^2F_t,\label{eq:convergence_pf_6}
\end{align}
where
\begin{align}
    &F_t=\frac{1}{C}\sum_{k\in\sK_t} s_k^2+6M\Gamma_t+8(E-1)^2G^2,\\
    &\Gamma_t=L_{\sK_t}^*-\frac{1}{C}\sum_{k\in\sK_t}l_k^*.
\end{align}
Eq.~\ref{eq:convergence_pf_6} arises from the inequality $1-Ex\le(1-x)^E\le 1-x $ for $x\in[0,1]$.

We now turn to bound the second term in Eq.~\ref{eq:convergence_pf_1}. We first find the $q_t$ in \cref{lemma7}.
\begin{align}
    \Arrowvert\vw^{t+1}-\vw^t\Arrowvert^2=&\Arrowvert\frac{1}{C}\sum_{k\in\sK_t}\vw^{t,E}_k-\vw^t\Arrowvert^2\\
    \le& \frac{1}{C}\sum_{k\in\sK_t}\Arrowvert\vw^{t,E}_k-\vw^t\Arrowvert^2\label{eq:convergence_pf_2}\\
    =&\frac{\eta_t^2}{C}\sum_{k\in\sK_t}\Arrowvert\sum_{i=0}^{E-1}\nabla l_k(\vw^{t,i}_k)\Arrowvert^2\\
    \le&\frac{\eta_t^2E}{C}\sum_{k\in\sK_t}\sum_{i=0}^{E-1}\Arrowvert\nabla l_k(\vw^{t,i}_k)\Arrowvert^2\label{eq:convergence_pf_3}\\
    \le&\frac{\eta_t^2E}{C}\sum_{k\in\sK_t}\sum_{i=0}^{E-1}G^2\label{eq:convergence_pf_4}\\
    \le&\frac{\eta_t^2E}{C}\sum_{k\in\sK_t}EG^2\\
    =&\eta_t^2E^2G^2=q_t^2,
\end{align}
where Eq.~\ref{eq:convergence_pf_2} and Eq.~\ref{eq:convergence_pf_3} comes from Jensen inequality, and Eq.~\ref{eq:convergence_pf_4} comes from Assumption~\ref{assumption_c4}. With Lemma \cref{lemma7}, we get
\begin{align}
    \Arrowvert\bm{\Sigma}^{t+1}-\bm{\Sigma}^t\Arrowvert_1\le &\frac{bN^2}{C}[G^2(\eta_t^2-\eta_{t+1}^2)+\\
    &2MEG\eta_t^3+M^2E^2G^2\eta_t^4)].
\end{align}
Further with a diminishing $\eta_t=\frac{\beta}{t+\gamma}$, we have
\begin{align}
    \eta_t^2-\eta_{t+1}^2=&\beta^2(\frac{1}{(t+\gamma)^2}-\frac{1}{(t+1+\gamma)^2})\\
    =&\beta^2\frac{2(t+\gamma)+1}{(t+\gamma)^2(t+1+\gamma)^2}\\
    \le&\frac{2\beta^2}{(t+\gamma)^3}\\
    =&\frac{2\eta_t^3}{\beta},
\end{align}
and with $\beta>\frac{1}{m}, \eta_t\le\eta_1\le \frac{1}{4M}$, we get 
\begin{align}
    &\Arrowvert\bm{\Sigma}^{t+1}-\bm{\Sigma}^t\Arrowvert_1\\
    \le&\frac{bN^2\eta_t^3}{C}(\frac{2G^2}{\beta}+2MEG+M^2E^2G^2\eta_t)\\
    \le& \frac{bN^2\eta_t^3}{C}(2mG^2+2MEG+\frac{1}{4}ME^2G^2)\\
    =&\eta_t^3 D,\label{eq:convergence_pf_5}
\end{align}
where
\begin{align}
    D=\frac{bN^2}{C}(2mG^2+2MEG+\frac{1}{4}ME^2G^2).
\end{align}
With Eq.~\ref{eq:convergence_pf_6} and Eq.~\ref{eq:convergence_pf_5}, we have
\begin{align}
    \Delta_{t+1}\le& (1-\eta_tm)\Delta_t+2E\eta_t^2F_t+(1+\frac{1}{\eta_t m})\eta_t^3\delta D\\
    \le&(1-\eta_tm)\Delta_t+\eta_t^2 (2EF_t+\frac{\delta}{m}D)+\eta_t^3 \delta D\\
    \le&(1-\eta_tm)\Delta_t+\eta_t^2(\tilde F+\tilde D)\label{eq:convergence_pf_7},
\end{align}
where 
\begin{align}
    \tilde F&=2E\max_{t}F_t,\\
    \tilde D&=(\frac{1}{m}+\frac{1}{4M})\delta D.
\end{align}

Now we can use the same trick in \cite{li2019convergence} to finish the proof of convergence. With a diminishing learning rate, $\eta_t=\frac{\beta}{t+\gamma}$ for some $\beta>\frac{1}{m}$ and $\gamma>0$ such that $\eta_1\le \min\{\frac{1}{m},\frac{1}{4M}\}=\frac{1}{4M}$, we will prove by induction that $\Delta_t\le\frac{\nu}{\gamma+t}$, where $\nu = \max\{\frac{\beta^2(\tilde F+\tilde D)}{\beta m-1},(\gamma+1)\Delta_1\}$. 

With the definition of $\nu$, we ensure that $\Delta_1\le\frac{\nu}{\gamma+1}$. Now we assume that $\Delta_t\le \frac{\nu}{\gamma+t}$ holds for some $t$, we have
\begin{align}
    \Delta_{t+1}\le&(1-\eta_tm)\Delta_t+\eta_t^2(\tilde F+\tilde D)\\
    \le&(1-\frac{\beta m}{t+\gamma})\frac{\nu}{t+\gamma}+\frac{\beta^2(\tilde F+\tilde D)}{(t+\gamma)^2}\\
    =&\frac{t+\gamma-1}{(t+\gamma)^2}\nu+\Big[\frac{\beta^2(\tilde F+\tilde D)}{(t+\gamma)^2}-\frac{\beta m-1}{(t+\gamma)^2}\nu\Big]\\
    \le&\frac{t+\gamma-1}{(t+\gamma-1)^2+2(t+\gamma)-1}\nu\label{eq:convergence_pf_8}\\
    \le&\frac{t+\gamma-1}{(t+\gamma-1)^2+2(t+\gamma-1)}\nu\\
    \le&\frac{\nu}{t+\gamma+1}.
\end{align}
Eq.~\ref{eq:convergence_pf_8} also arises from the definition of $\nu$ that $\beta^2(\tilde F+\tilde D)\le (\beta m-1)\nu$. Accordingly, for all $t$, we have $\Delta_t\le \frac{\nu}{\gamma+t}$ holds. 
\end{proof}
\end{theorem}

With this result, we prove that $\Delta_t$ converges to $0$ with convergence rate $\mathcal{O}({\frac{1}{T}})$, and thus we can say that the proxy algorithm of FedCor converges to the global optimal with convergence rate $\mathcal{O}({\frac{1}{T}})$ with Corollary~\ref{corollary3}. 

\section{Experiment Details}
We simulate the training process of federated learning on one machine. All experiments in this paper are run on one NVIDIA 2080-Ti GPU and two Intel Xeon E5-2630 v4 CPUs. The experiments on FMNIST require around 3 hours for each seed, and the experiments on CIFAR-10 require around 10 hours for each seed.
\subsection{Model Parameters}\label{Apx:Model}
\paragraph{Hyperparameters in FMNIST} We follow \cite{cho2020client} to construct the neural model on FMNIST: An MLP model with two hidden layers with $64$ and $30$ units, respectively. Under all three heterogeneous settings, we set the local batch size $B=64$ and the number of local iterations $E=20$. The learning rate $\eta_0$ is set to $0.005$ initially, and halved at the 150-th and 300-th rounds. An SGD optimizer with a weight decay of $0.0001$ and no momentum is used. We allocate data to $N=100$ clients, and set the participation fraction $C=10$ for the 1SPC setting, and $C=5$ for the 2SPC and Dir settings.
\paragraph{Hyperparameters in CIFAR-10} We use a CNN with three convolutional layers~\cite{tensorflow2016tensorflow} with $32$, $64$ and $64$ kernels, respectively. And all convolution kernels are of size $3\times 3$. Finally, the outputs of convolutional layers are fed into a fully-connected layer with 64 units. Under all three heterogeneous settings, we set the local batch size $B=50$ and the number of local iterations $E=40$. We use a learning rate $\eta=0.01$ without learning rate decay, and a weight decay of $0.0003$ for the SGD optimizer. The total number of clients and the client participation fraction are the same as those in FMNIST.
\paragraph{Hyperparameters for FedCor} We set the dimension of client embedding $d=15$ for all experiments. In \cref{eq:train}, we set $M=10,S=1$ for the warm-up phase, and $M=1,S=1$ for the normal phase. And we set the discount factor $\gamma=\theta^{\Delta t}$ where $\theta = 0.9$ for experiments on FMNIST and $\theta=0.99$ for experiments on CIFAR-10. In each GP update round $t$, we use $\mX^{t-1}$ as the initialization and use an Adam optimizer~\cite{kingma2014adam} with learning rate $0.01$ to optimize for $\mX^t$. Notice that although \cref{eq:train} has a closed form optimal solution for $\mX^t$, we still learn $\mX^t$ with the gradient decent method with the initialization $\mX^{t-1}$ in order to utilize the covariance stationarity and reduce the evaluation bias with small number of samples.
\paragraph{Hyperparameters for other baselines}
We use the same parameters $\alpha_1=0.75,\alpha_2=0.01$ and $\alpha_3=0.1$ as those in the paper~\cite{goetz2019active} for Active Federated Learning. And we set $d=2NC$ for Power-of-choice Selection Strategy, which is empirically shown to be the best value of $d$ in a highly heterogeneous setting in the paper~\cite{cho2020client}.

Note that we implement the random selection strategy as uniformly sampling clients from $\sU$ without replacement~\cite{mcmahan2017communication}, while Cho et al.\cite{cho2020client} implement the random selection strategy as sampling clients with replacement. Thus, our implemented random selection strategy achieves better performances than their implementation. 
\begin{figure*}[ht]
\centering
\includegraphics[width=0.75\linewidth]{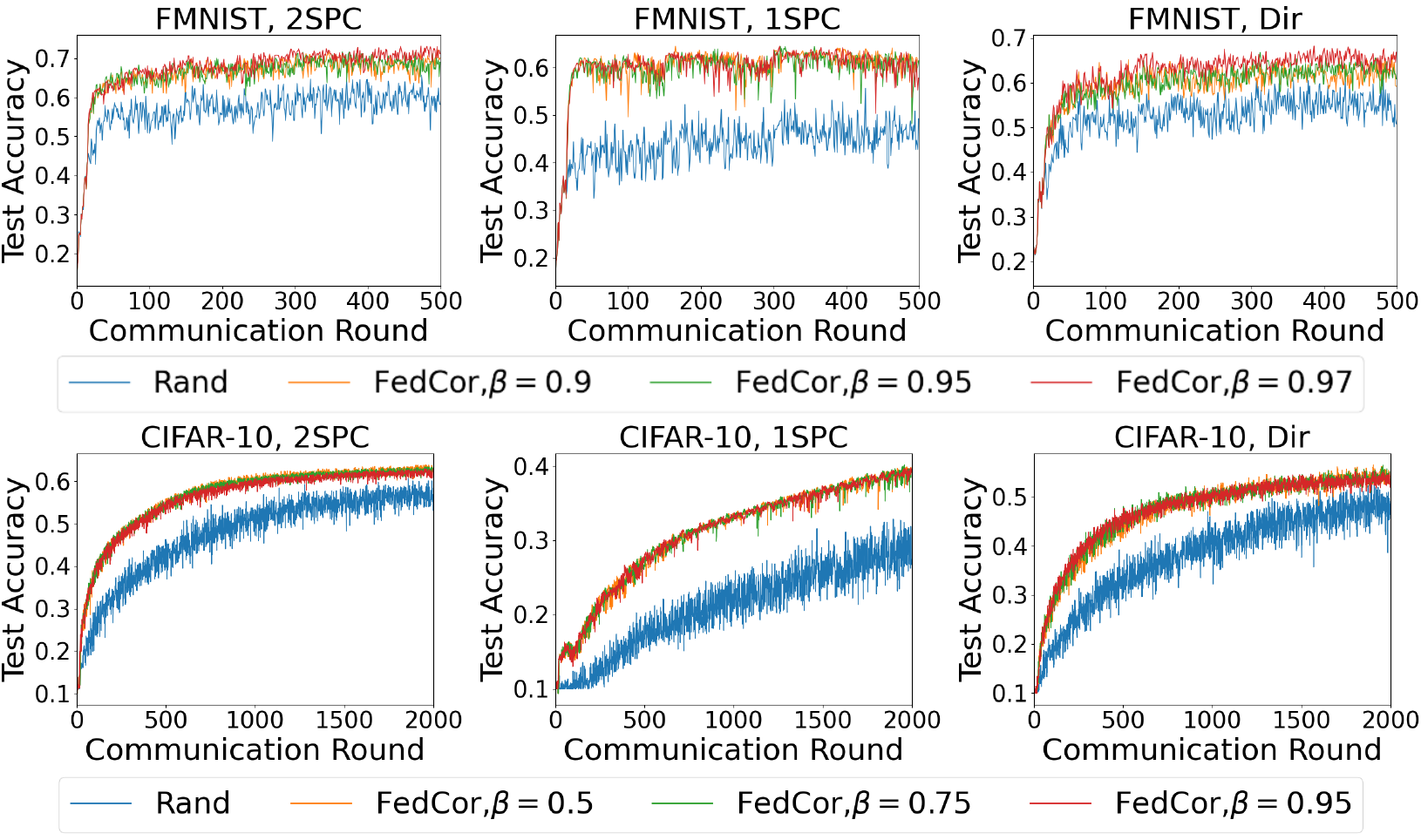}
\caption{Test accuracy with different annealing coefficient $\beta$ on FMNIST (top) and CIFAR-10 (bottom) under three heterogeneous settings (left: 2SPC; median: 1SPC; right: Dir).}
\label{fig:anneal_acc}
\end{figure*}
\subsection{Dirichlet Distribution for Data Partition}\label{Apx:Dirichlet}
We follow the idea in~\cite{hsu2019measuring} to construct the Dir heterogeneous setting, while we make some modifications to get an unbalanced non-identical data distribution.

For each client $k$, we sample the data distribution $\boldsymbol q_k\in \mathbb{R}^{10}$ from a dirichlet distribution independently, which could be formulated as
\begin{equation}
    \boldsymbol q_k \sim \text{Dir}(\alpha \boldsymbol p),
\end{equation}
where $\boldsymbol p$ is the prior label distribution and $\alpha\in\mathbb{R}_+$ is the concentration parameter of the dirichlet distribution. We group $\boldsymbol q_k$ of all the clients together and get a fraction matrix $Q=[\boldsymbol q_1,\cdots, \boldsymbol q_n]$. We denote the size of dataset on each client as $\boldsymbol x=[x_1,\cdots,x_N]^T$ and we get it from a solution of a quadratic programming: 
\begin{align}
    \min_{\boldsymbol x}\quad&\boldsymbol x^T\boldsymbol x\\
    \text{subject to }\quad&Q\boldsymbol x=\boldsymbol d\\
    &\boldsymbol x\in \mathbb{R}_{++}^N,
\end{align}
where $\boldsymbol d$ is the number of data with each label. We minimize $\Vert\boldsymbol x\Vert_2$ to avoid the cases where data distribution is over-concentrated on a small fraction of clients. In that case, the client selection problem might become trivial, since we can always ignore those clients with a small dataset and select those with a large dataset.

\section{Extra Experimental Results}
\subsection{Ablation Study: Annealing Coefficient}\label{Apx:beta}

We conduct experiments on FMNIST and CIFAR-10 with different annealing coefficient $\beta$. We setup our experiments under three heterogeneous settings as in Section~\ref{Sec:Exp}, with different annealing coefficient $\beta$ ($\beta=0.95,0.75,0.5$ for FMNIST and $\beta=0.97,0.95,0.9$ for CIFAR-10). We fix the GP training interval $\Delta t$ to $10$ for FMNIST and $50$ for CIFAR-10. The test accuracy curves are shown in Figure~\ref{fig:anneal_acc}. We can see that within a large range, the value of annealing coefficient only slightly influence the convergence rate as well as the final accuracy. Recalling the results of different GP training intervals $\Delta t$ in Section~\ref{SubSec:Int}, we can say that our method is not sensitive to the hyperparameters $\Delta t$ and $\beta$.

\begin{figure*}[ht]
\centering
\includegraphics[width=0.96\linewidth]{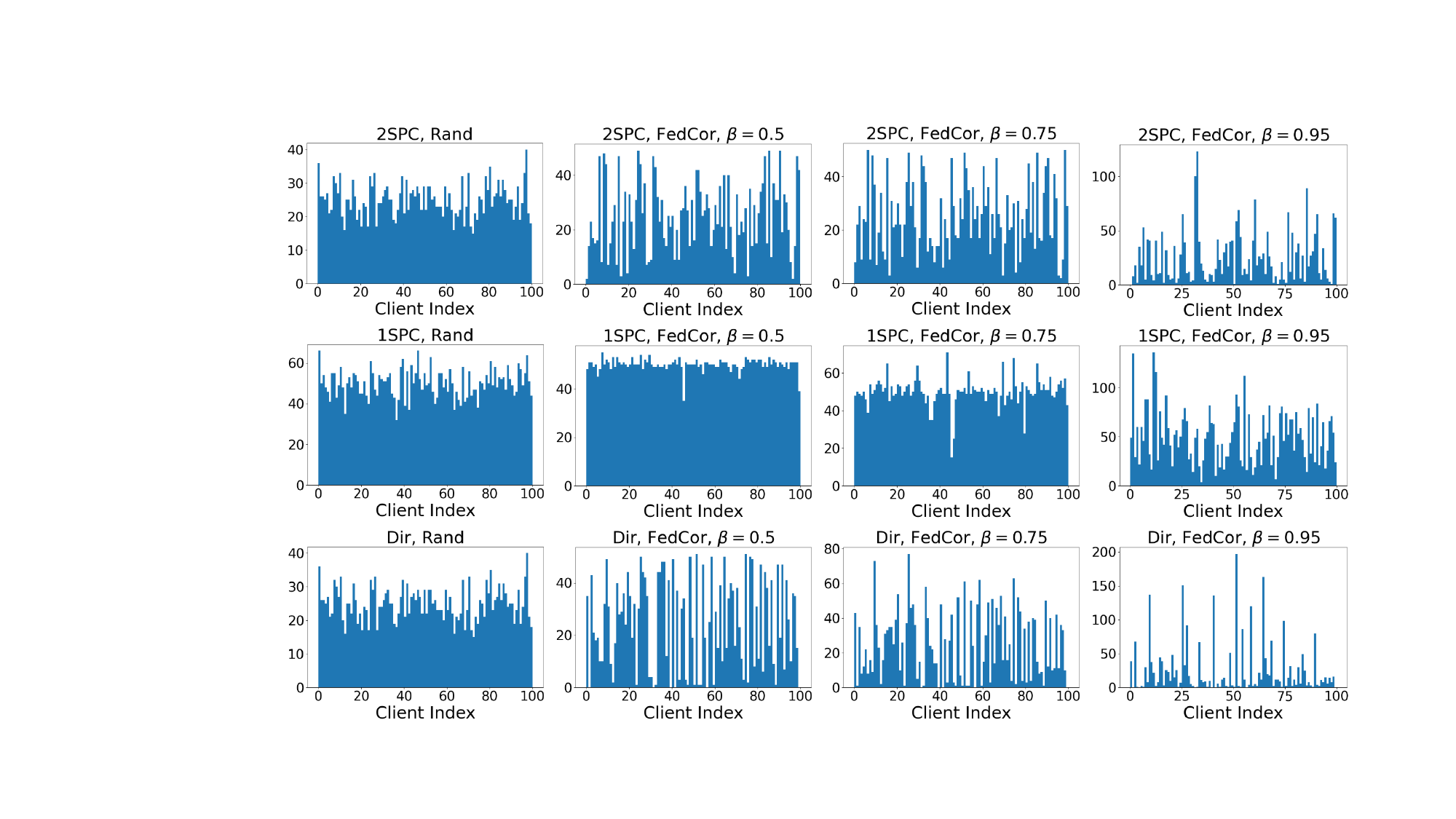}
\caption{Selected Frequency of each client with different annealing coefficient $\beta$ on FMNIST under three heterogeneous settings (top: 2SPC; median: 1SPC; bottom: Dir).}
\label{fig:fmnist_hist}
\end{figure*}

\begin{figure*}[ht]
\centering
\includegraphics[width=0.96\linewidth]{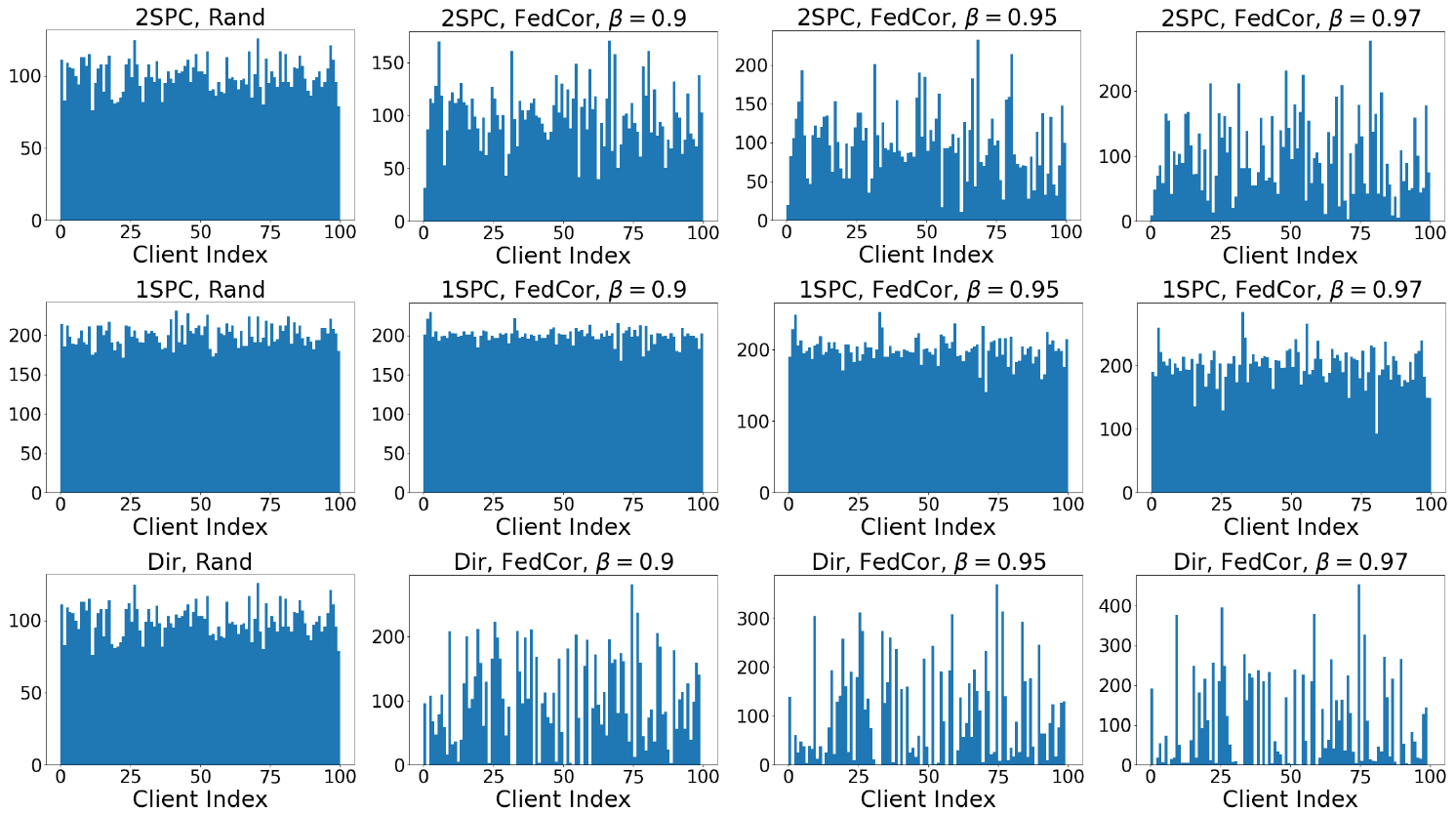}
\caption{Selected Frequency of each client with different annealing coefficient $\beta$ on CIFAR-10 under three heterogeneous settings (top: 2SPC; median: 1SPC; bottom: Dir).}
\label{fig:cifar_hist}
\end{figure*}

We present the selected frequency of each client in Figure~\ref{fig:fmnist_hist} and Figure~\ref{fig:cifar_hist} for FMNIST and CIFAR-10 respectively. We can see that with a smaller $\beta$, the selected frequency tends to be more ``uniform''. However, this does not mean that our selection strategy is equivalent to the uniformly random selection. Our sequential selection strategy introduces dependencies between selected clients as discussed in the multi-iteration insights in Section~\ref{SubSec:Sel}, which makes our selection strategy prefer some combinations of selected clients to others, while the uniformly random selection treats all the combinations equally. The advantage shown in Figure~\ref{fig:anneal_acc} compared to the uniformly random strategy demonstrates that selecting a good combination of clients, not only a good individual, is important.
\begin{figure*}[t]
\centering
\includegraphics[width=0.6\linewidth]{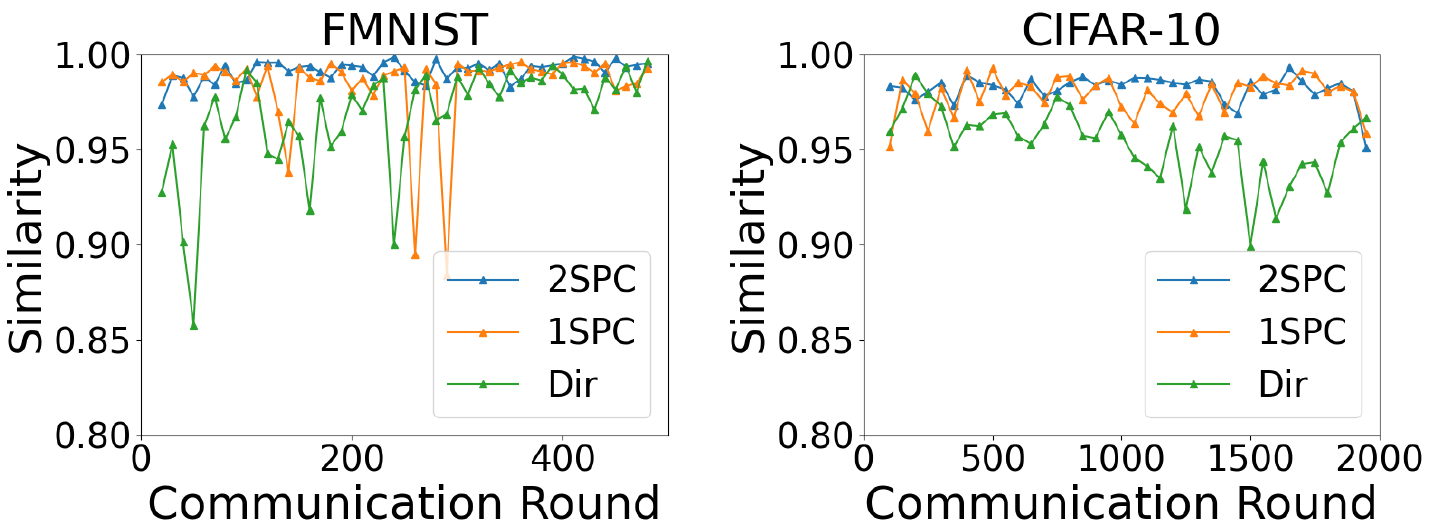}
\caption{Verification of covariance stationarity on FMNIST and CIFAR-10.}
\label{fig:cov_stat}
\end{figure*}
\subsection{Normality Verification}\label{Apx:mvnt}

We setup experiments to show that Gaussian Distribution can model the loss changes w.r.t. uniformly sampled client selection. To verify this, in the last round of the warm-up phase, we perform the following procedure to examine the normality.
\begin{enumerate}
    \item We uniformly sample $1000$ different client selections $\{\sS_{t,i}:i=1,\cdots,1000\}$ and collect the corresponding loss changes $\Delta\vl^t(\sS_{t,i})=[\Delta l_1^t(\sS_{t,i}),\cdots,\Delta l_N^t(\sS_{t,i})]$ for each of them.
    \item We perform PCA on $\{\Delta\vl^t(\sS_{t,i}):i=1,\cdots,1000\}$ to extract the principle components.
    \item We plot the histogram of each principle component and compare its distribution with the Gaussian Distribution.
\end{enumerate}
We do not use Multivariate Normality Test directly because we find that $\bm{\Sigma^t}$ is always nearly singular, which makes the Multivariate Normality Test unstable. Thus, we turn to perform PCA and visualize each principle component to verify the normality.

\begin{figure*}[tb]
\centering
\includegraphics[width=0.8\linewidth]{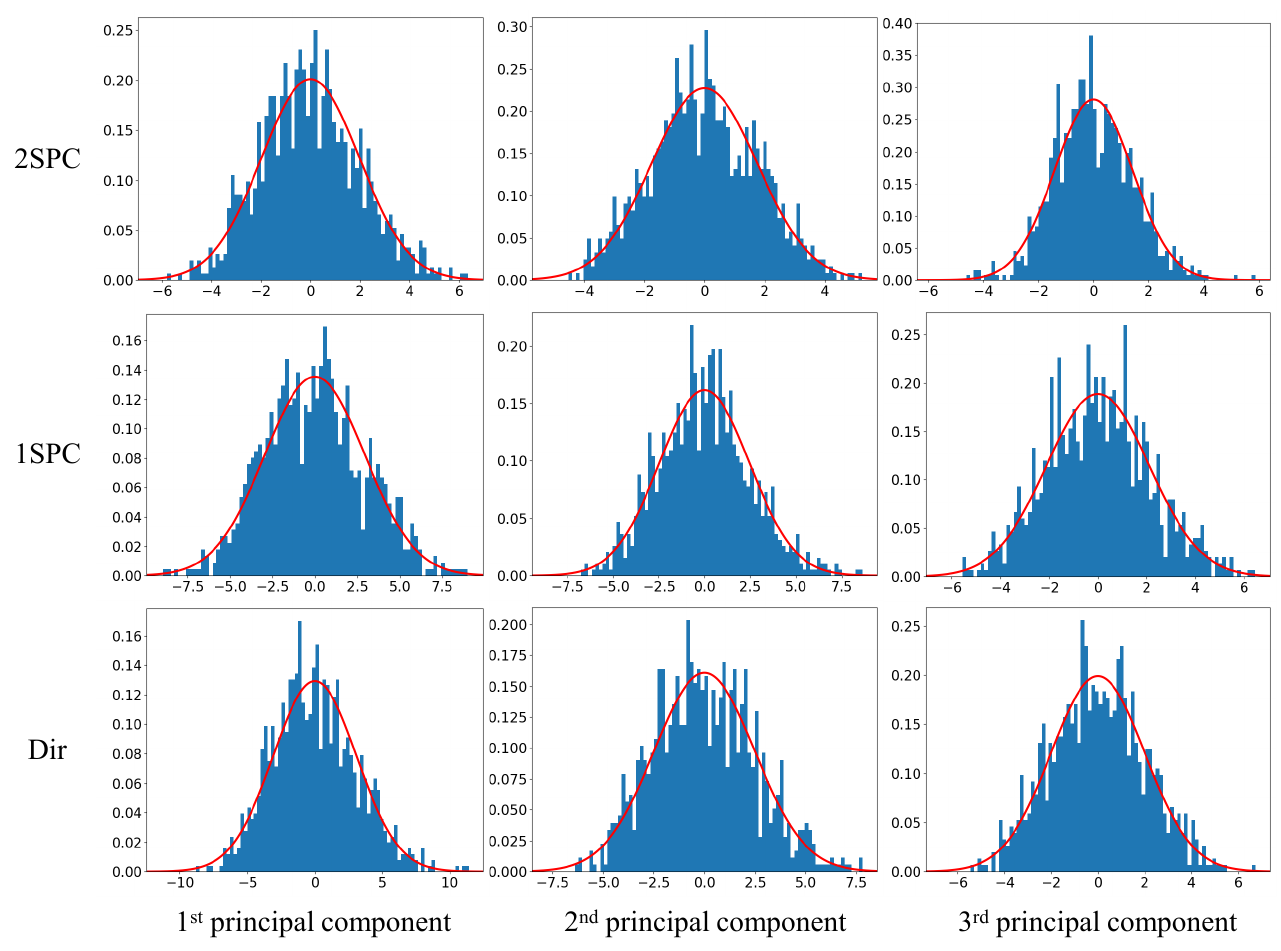}
\vspace{-0.5em}
\caption{Normality Test on FMNIST.}
\label{fig:fmnist_mvnt}
\end{figure*}

\begin{figure*}[tb]
\centering
\includegraphics[width=0.8\linewidth]{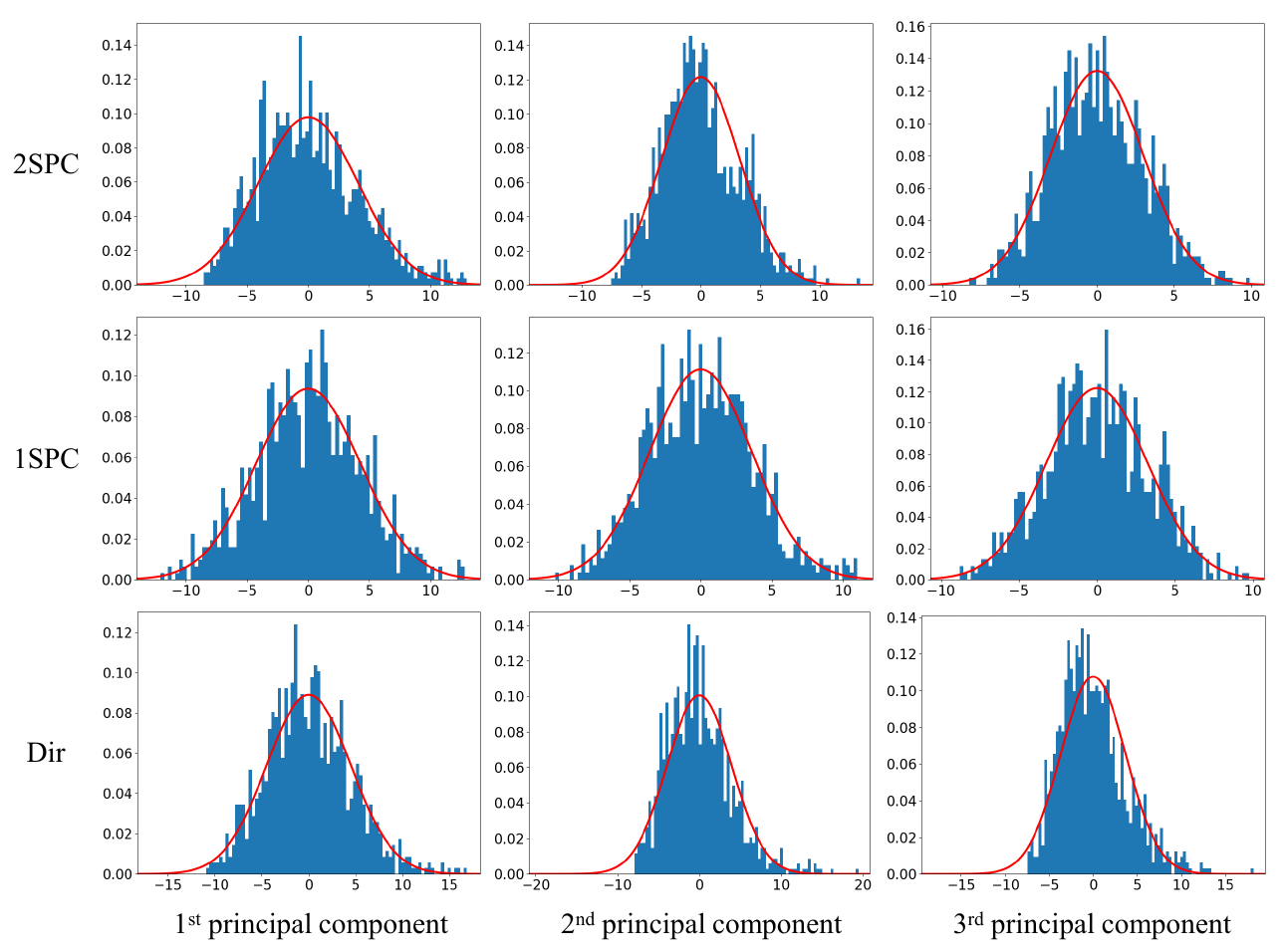}
\vspace{-0.5em}
\caption{Normality Test on CIFAR-10.}
\label{fig:cifar_mvnt}
\end{figure*}

The results of FMNIST and CIFAR-10 are shown in Figure~\ref{fig:fmnist_mvnt} and Figure~\ref{fig:cifar_mvnt} respectively. The red line shows the probability density of Gaussian Distribution with the mean and variance of that principle component. We can see that in all our experiments, Gaussian Distribution can fit the distribution of the principle component well, which verifies that Lemma~\ref{lemma1} does hold in all the experiment settings.

\subsection{Covariance Stationarity Verification}\label{Apx:cov_stat}
We examine that assumption in Section~\ref{SubSec:Train} that the covariance keep approximately stationary during the FL training, namely,
\begin{equation}
\forall t,{\bm{\Sigma}}^t\approx {\bm{\Sigma}}^{t+\Delta t}.
\end{equation}
To verify this, every $\Delta t$ rounds ($\Delta t=10$ for FMNIST and $\Delta t=50$ for CIFAR-10), we randomly sample $1000$ client selections $\sK_i$ and collect the corresponding loss changes $\Delta\vl^t(\sK_i)$. We directly calculate the covariance matrix ${\bm{\Sigma}}^t$ with these samples $\{\Delta\vl^t(\sK_i):i=1,\cdots,1000\}$. Then for each adjacent pair of covariance matrix, we calculate their cosine similarity as follows.
\begin{equation}
\rm{similarity}({\bm{\Sigma}}^t,{\bm{\Sigma}}^{t+\Delta t})=\frac{\rm{tr}({{\bm{\Sigma}}^t}^T{\bm{\Sigma}}^{t+\Delta t})}{\rm{tr}({{\bm{\Sigma}}^t}^T{\bm{\Sigma}}^{t})\rm{tr}({{\bm{\Sigma}}^{t+\Delta t}}^T{\bm{\Sigma}}^{t+\Delta t})}
\end{equation}
The similarity is in range $[0,1]$, and a larger one shows a higher similarity.

The results are shown in Figure~\ref{fig:cov_stat}. We can see that in most cases the similarity is larger than $0.9$, which verifies our claim of the covariance stationarity.

\end{document}